%% file: main.tex
\documentclass[letterpaper, 10 pt, journal, twoside]{IEEEtran}  

\usepackage{optidef}
\usepackage{dsfont}
\usepackage{booktabs}
\usepackage{multirow}
\usepackage{siunitx}
\usepackage{makecell} 
\usepackage{ral}
\usepackage{fontawesome5}

\acrodef{MPC}{model predictive control}
\acrodef{RL}{Reinforcement Learning}
\acrodef{UAV}{Unmanned Aerial Vehicle}
\acrodef{CTBR}{collective thrust and body rates}
\acrodef{PPO}{Proximal Policy Optimization}
\acrodef{HIL}{Hardware-in-the-Loop}

\newcommand{\reb}[2][\relax]{%
  \ifx#1\relax
    \textcolor{blue}{#2}%
  \else
    \textcolor{blue}{[#1]: #2}%
  \fi
}
\renewcommand{\reb}[2][]{#2}
\newcommand{\rev}[1]{#1}

\title{Learning to Throw: Agile and Accurate Cable-Suspended \\Payload Delivery with a Quadrotor}
\author{Yifan Zhai, Elia Raimondi, Yunfan Ren, Ismail Geles, Yannick Armati, Jiaxu Xing, and Davide Scaramuzza
\thanks{Manuscript received: June 25, 2026; Revised September 20, 2026; Accepted November 23, 2026. This paper was recommended for publication by
Editor Giuseppe Loianno upon evaluation of the Associate Editor and Reviewers' comments. }
\thanks{
Supported by the European Union’s Horizon Europe programme (grant No. 101120732, AUTOASSESS) and the European Research Council (ERC)
(grant No. 864042, AGILEFLIGHT). All authors are with the Robotics and Perception Group, Department of Informatics, University of Zurich, Switzerland (\protect\url{https://rpg.ifi.uzh.ch}).
        {Contact: \tt\small dzhai@ifi.uzh.ch}}%
}

\begin{document}

\maketitle

\input{sections/abstract.tex}


\acresetall 

\input{sections/supplementary_materials}

\input{sections/intro.tex}
\input{sections/related_work.tex}
\input{sections/method.tex}
\input{sections/experiments}


\acresetall 

\input{sections/conclusion.tex}

\bibliographystyle{IEEEtran}
\bibliography{reference}

\vfill

\end{document}

%% file: sections/abstract.tex
\begin{abstract}
While quadrotors have proven effective for rapidly transporting suspended payloads in time-critical applications, 
the dynamic, targeted release of these payloads remains largely underexplored.
%
State-of-the-art approaches typically rely on model-based trajectory optimization and tracking; however, these methods often yield sub-optimal performance due to simplified dynamics models, tracking errors, and conservative feasibility constraints.
%
%
\reb{In this work, instead of planning trajectories on a simplified model, we directly train a throwing policy on the full quadrotor-payload dynamics. To this end, we propose a hybrid simulator that separates quadrotor force generation from dynamics simulation: a system-identified quadrotor simulator supplies the body wrench to a multibody solver that advances the quadrotor, rope, and payload together.} 
%
\reb{The reinforcement learning policy trained in this simulator commands the quadrotor directly, bypassing the intermediate trajectory representation, and thereby uses more of the platform's authority to execute agile and accurate throws.}
%
\reb{Deployed zero-shot on hardware and compared against two state-of-the-art methods, our policy lands the payload within \qtyrange{7.9}{11.2}{\centi\meter} of targets \qtyrange{1.5}{2.5}{\meter} away on average; it reduces the landing error by $69$--$74\%$ and the throw duration by $33$--$47\%$ relative to a trajectory-optimization baseline, and throws $2.2$--$2.5\times$ faster than an error-bounded delivery planner.}
\rev{Ablation studies further show that the policy's accuracy is enabled by the modeled simulation components.}
\reb{We further demonstrate accurate visual feature-based throwing in hardware-in-the-loop experiments.}
To accelerate research in dynamic aerial manipulation, we open-source the simulator.

\end{abstract}

%% file: sections/supplementary_materials.tex
\hypersetup{
  colorlinks=true,
  urlcolor=ethblue
}

\vspace{-0.2cm}
\section*{Supplementary Materials}

\noindent
\faVideo\hspace{0.6em}\text{Video:}
\url{https://youtu.be/Psq-tH6NuFQ}

\noindent
\faGlobe\hspace{0.6em}\text{Website:}
\url{https://rpg.ifi.uzh.ch/learn_to_throw/}

\vspace{-0.2cm}

%% file: sections/intro.tex
\section{Introduction}

\IEEEPARstart{Q}{uadrotors} carrying a cable-suspended payload can provide fast, flexible aerial transportation.
In time-critical missions such as search-and-rescue and disaster relief, accurately throwing the payload at its target not only enables faster delivery, but also conserves energy and, in turn, extends mission range.
%
However, agile and accurate throwing is hard: the quadrotor and its load form a highly nonlinear system in which the payload is actuated only indirectly, and after release, the payload follows an uncontrolled ballistic trajectory, so small errors at release become large landing errors.

\reb{Most existing payload delivery methods are model-based, optimizing a trajectory that is tracked by a separate controller~\cite{Foehn_Falanga_Kuppuswamy_Tedrake_Scaramuzza_2017, Cao_Fang_Liang_2025, AD_planner}. To make the optimization tractable, they adopt simplified dynamics models such as idealized quadrotor actuation~\cite{Foehn_Falanga_Kuppuswamy_Tedrake_Scaramuzza_2017, pcmpc2021} and an always-taut cable~\cite{AD_planner, pcmpc2021}. This model mismatch causes the release state to deviate from the planned one, which translates directly into landing error~\cite{Foehn_Falanga_Kuppuswamy_Tedrake_Scaramuzza_2017}. Bounding that error requires conservative release constraints, which keep the delivery accurate but sacrifice the agility of a throw~\cite{AD_planner}.}

\begin{figure}[!t]
\centering
\includegraphics[width=\columnwidth]{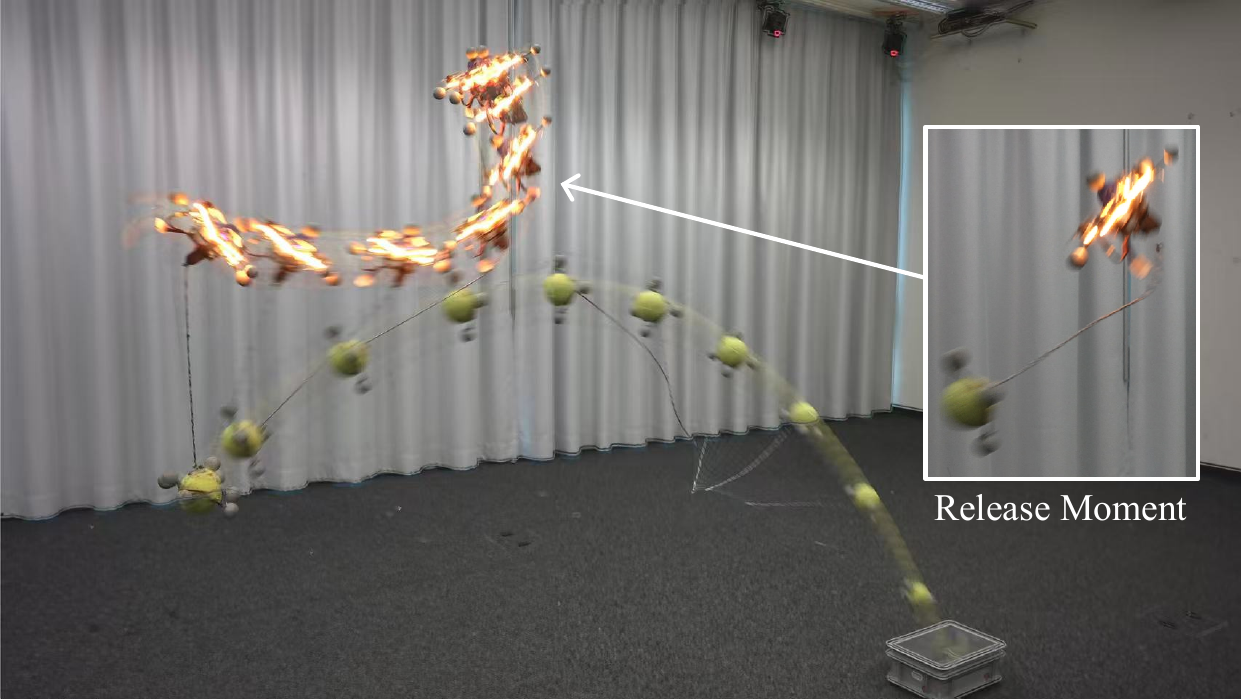}\\[2mm]
\begin{tikzpicture}[font=\footnotesize, >={Latex[length=1.8mm]}]
  \node[draw=blue!45, fill=blue!6, rounded corners=3pt, inner sep=1.2mm, align=center] (ana)
    {\includegraphics[width=22mm]{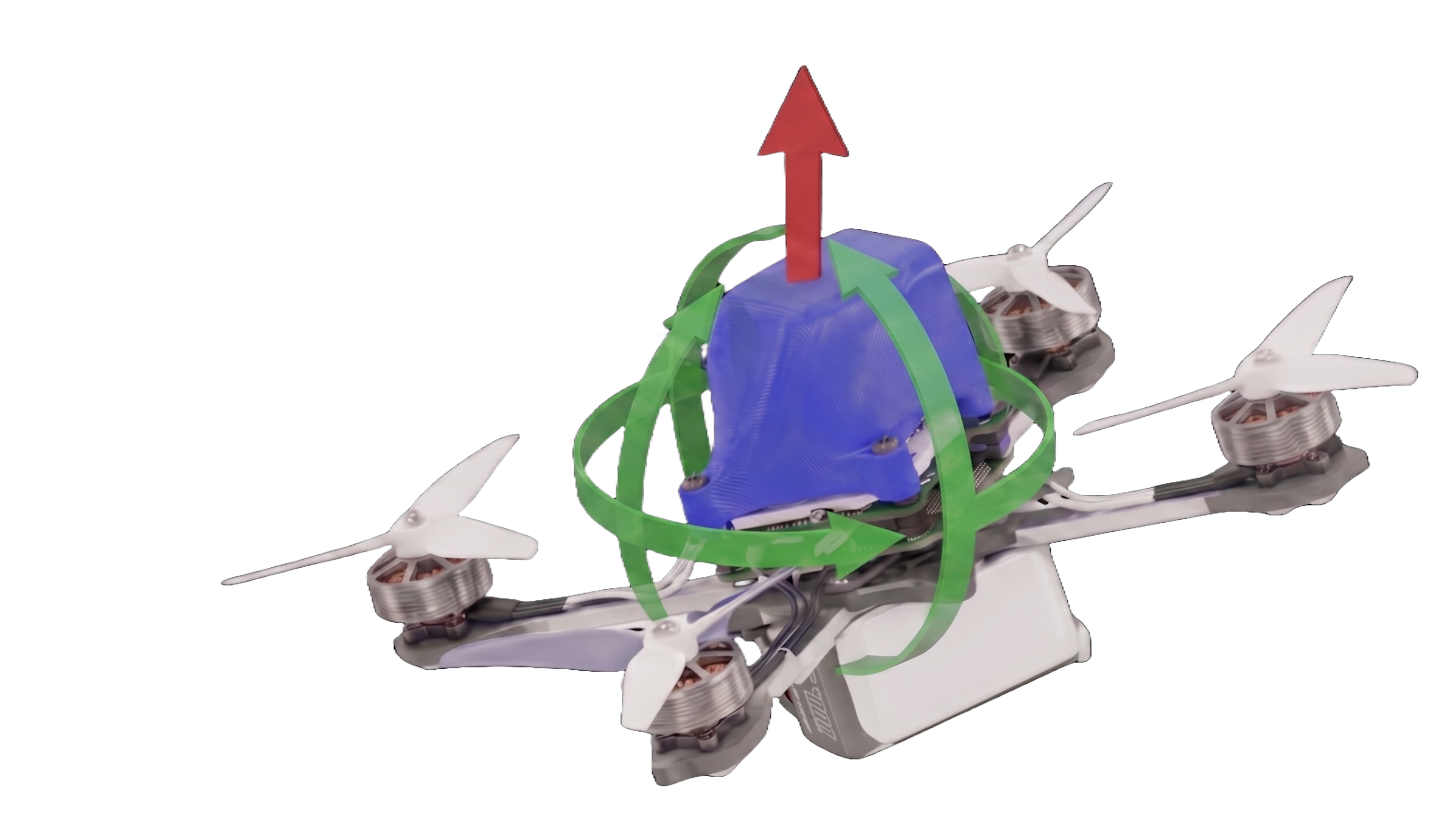}\\[0.2mm]
     \textbf{\textcolor{blue!55}{Standalone Quadrotor}}\\
     \textbf{\textcolor{blue!55}{Simulator}}};
  \node[draw=orange!55, fill=orange!7, rounded corners=3pt, inner sep=1.2mm, align=center,
        right=28mm of ana] (px)
    {\includegraphics[width=15mm]{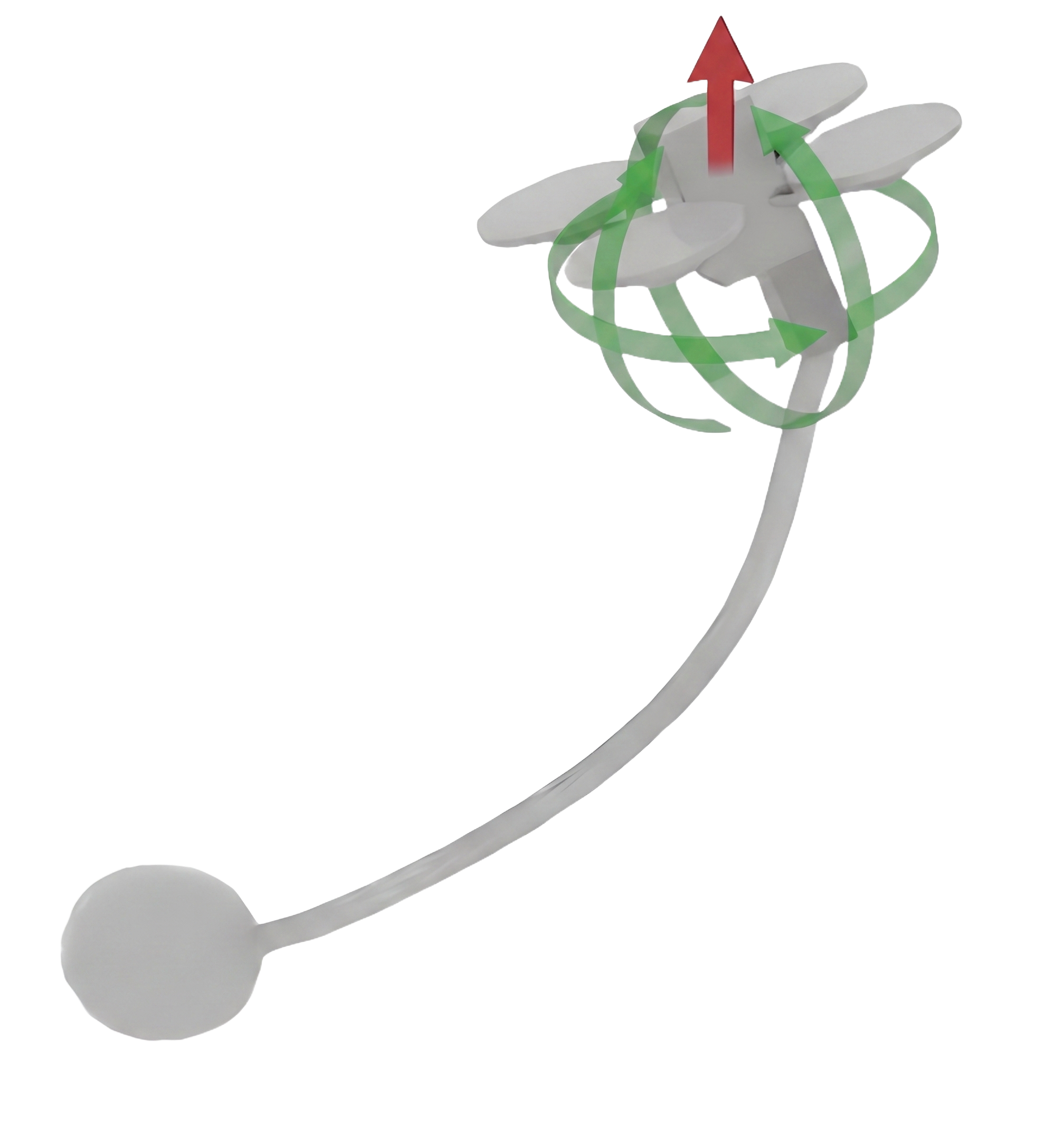}\\[0.2mm]
     \textbf{\textcolor{orange!60}{Rope \& payload}}\\
     \textbf{\textcolor{orange!60}{Simulator}}};
  \draw[->, semithick] ([yshift=5mm]ana.east) --
     node[above, font=\scriptsize, inner sep=1.5pt] {Quad body $\mathbf{F}_b,\boldsymbol{\tau}_b$}
     ([yshift=5mm]px.west);
  \draw[->, semithick] ([yshift=-5mm]px.west) --
     node[below, font=\scriptsize, inner sep=1.5pt] {Quad body $\mathbf{p},\mathbf{q},\mathbf{v},\boldsymbol{\omega}$}
     ([yshift=-5mm]ana.east);
\end{tikzpicture}
\caption{\emph{Top:} Zero-shot deployment of our RL policy. \emph{Bottom:} The hybrid simulator used to train the policy. It couples a quadrotor simulator, modeled on real standalone quadrotor flight data, with an external physics solver.}
\label{fig:teaser}
\vspace{-0.4cm}
\end{figure}

\reb{\ac{RL} offers an alternative: training on the full dynamics model minimizes model mismatch, while directly outputting vehicle commands avoids the intermediate trajectory representation. This approach has enabled agile transport and traversal with attached payloads~\cite{zeng2025decentralizedaerialmanipulationcablesuspended, flare2026}, though with simplified actuator and controller dynamics in simulation~\cite{flare2026} or with the remaining mismatch left to disturbance rejection on the vehicle~\cite{zeng2025decentralizedaerialmanipulationcablesuspended}. In these tasks, where the payload remains attached, the policy can still correct the resulting errors during the maneuver. Throwing the payload, however, allows no correction after release, so the policy must learn precisely how vehicle actuation generates forces and how the cable transmits them to build payload momentum.}

\reb{To address this challenge, we propose a hybrid simulator that separates vehicle force generation from coupled mechanical integration, as shown in Fig.~\ref{fig:teaser}. A standalone quadrotor simulator, identified from real flights, advances its own motor, controller, and battery states and supplies the body wrench to a multibody solver that integrates the quadrotor, rope, and payload together, so the identified quadrotor response to \ac{CTBR} commands is preserved while cable reactions are resolved.}

\reb{Using this simulator, we train an \ac{RL} policy to throw the payload at designated targets. Deployed zero-shot on hardware against two state-of-the-art model-based methods~\cite{Foehn_Falanga_Kuppuswamy_Tedrake_Scaramuzza_2017, AD_planner}, it throws with high accuracy and substantially faster than both}; \rev{ablation studies show that each tested simulator component is needed for accurate transfer}.
\reb{We further extend the framework to visual feature-based throwing, evaluated in a hardware-in-the-loop setup with virtual observations~\cite{xing2024bootstrapping, heegIcra2025, pan2025learning}.}

\textit{Contributions:} \reb{We propose a hybrid simulator that preserves the identified quadrotor response within the coupled cable--payload dynamics, enabling agile and accurate throwing through direct \ac{CTBR} commands and zero-shot hardware transfer.}
\reb{We conduct hardware experiments against two model-based baselines, ablations, and a visual feature-based extension to support these claims.}
%

%% file: sections/related_work.tex
\section{Related Work}

\subsection{Model-Based Control of Suspended-Payload Systems}
\label{subsec:rw_model_based}
A large body of model-based work has studied quadrotors carrying cable-suspended payloads, whose dynamics switch between force-transmitting \emph{taut} modes and free-falling \emph{slack} modes.
The geometric-control formulation of Sreenath~\textit{et al.}~\cite{6760219} models the system on $SE(3)\times S^2$ with a rigid, massless cable and a point-mass load, establishes differential flatness, and treats taut/slack transitions as hybrid switches.
Subsequent work builds on similar abstractions for event-camera-based cable estimation~\cite{panetsos2023event}, cluttered-environment transport~\cite{li2023autotrans}, aggressive gap traversal~\cite{10478625}, and cooperative multi-quadrotor lifting~\cite{sun2025agilecooperative}. \reb{To keep planning and control tractable, such methods simplify both the vehicle and the cable.}
\reb{For example, perception-constrained \ac{MPC}~\cite{pcmpc2021, recalde2025eshpc} treats collective thrust and body rates as direct inputs of its prediction model and enforces positive cable tension, omitting actuator transients and slack motion, while RotorTM~\cite{10328685} models taut/slack transitions and impacts but takes thrust and moment as inputs and omits motor dynamics. Richer cable models exist, such as deformable continuum formulations~\cite{rapuano2026nonlinear}, but they have so far been validated only in simulation.}
\reb{The recent AD-Planner~\cite{AD_planner} extends model-based planning to the full pickup, in-flight delivery, and drop-off pipeline, by casting the payload under error-bounded release constraints that certify the landing accuracy. It plans through differential flatness on the same taut, massless-cable, point-mass model, and enforces positive cable tension, so slack-cable motion and the vehicle's actuation response are not considered.}
\vspace{-0.3cm}
\subsection{Learning-Based Suspended-Payload Flight}
Learning-based methods have recently been used to avoid hand-designed models of coupled rope-payload dynamics.
Early work applies model-based meta-\ac{RL} to adapt online to unknown payload mass and cable length, but remains limited to slow transport and drop-off tasks~\cite{Belkhale_2021}.
\reb{Recent works train policies for decentralized cooperative transport~\cite{zeng2025decentralizedaerialmanipulationcablesuspended} and agile single-quadrotor flight with an attached payload~\cite{flare2026}.}
\reb{These works keep the payload attached throughout the maneuver, and they differ in how the vehicle itself is simulated and in how the remaining sim-to-real gap is closed. Zeng~\textit{et al.}~\cite{zeng2025decentralizedaerialmanipulationcablesuspended} report that \ac{CTBR} policies do not learn effectively in their cooperative task, and deploy acceleration and body-rate commands tracked by an INDI-based controller that rejects cable disturbances. OmniDrones~\cite{xu2024omnidrones} models the rotor response, but its payload tasks use a rigid link and are evaluated only in simulation. FLARE~\cite{flare2026} integrates the coupled quadrotor--cable--payload mechanics in a single engine, with a memoryless body-rate controller and without motor-response or battery states, actuation delay, or aerodynamic drag.}
\reb{None of these learning-based methods addresses the dynamic release that defines a throw. Our simulator preserves the identified quadrotor response within articulated cable dynamics, enabling direct \ac{CTBR} throwing on hardware.}

\vspace{-0.3cm}
\subsection{Aerial Throwing and Targeted Release}
\reb{To the best of our knowledge, all existing solutions for throwing a cable-suspended payload are model-based: they optimize a release state using the reduced vehicle--payload models discussed in Sec.~\ref{subsec:rw_model_based}. Foehn~\textit{et al.}~\cite{Foehn_Falanga_Kuppuswamy_Tedrake_Scaramuzza_2017} use rotor thrusts as direct inputs, without modeling motor response or the onboard controller, and treat the payload as a point mass whose complementarity constraints permit taut/slack transitions. Cao~\textit{et al.}~\cite{Cao_Fang_Liang_2025} accelerate time-optimal planning with an analytic initial guess. Both produce nominal trajectories executed by a separate controller, so any mismatch between the modeled and realized response shifts the release state, and the ballistic flight amplifies it into landing error. AD-Planner~\cite{AD_planner} accounts for this uncertainty by propagating bounded release-state errors into a worst-case landing-error constraint. Since this bound grows with the payload's flight time, its horizontal speed, and its acceleration at release, it favors low, slow, and gentle casts, trading agility for accuracy.}

%% file: sections/method.tex
\section{Methodology}
%


\subsection{Quadrotor Hardware and Release Mechanism}
\label{subsec:hardware}
\begin{figure}[!t]
\centering
\includegraphics[width=\linewidth]{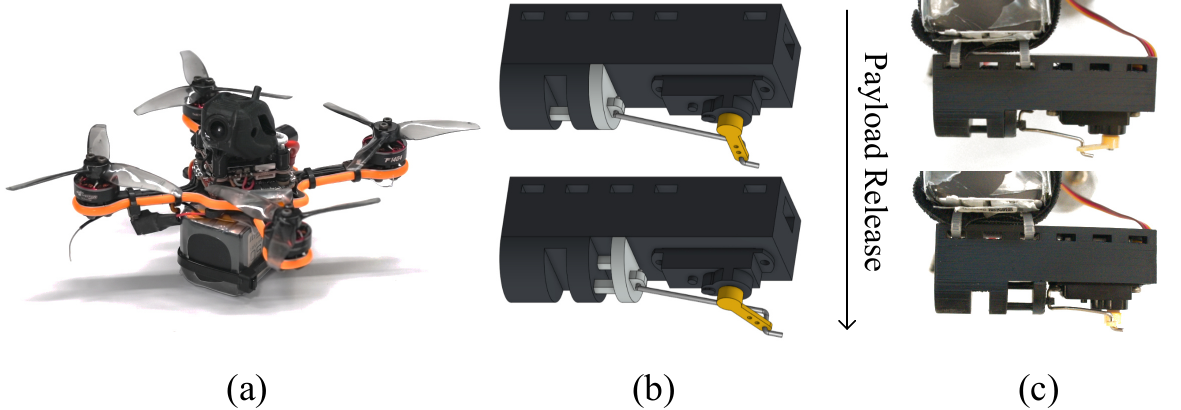}
\caption{Hardware used: (a) the standalone quadrotor, (b) the release mechanism CAD model, and (c) the 3D-printed release mechanism secured to the standalone quadrotor battery strap in its closed (top) and open (bottom) states.}
\label{fig:hardware}
\vspace{-0.4cm}
\end{figure}

As shown in Fig. \ref{fig:hardware}, the custom quadrotor used has a mass of \qty{0.21}{\kg} and arm length $\qty{19.4}{\cm}$, with propeller radius of \qty{3.81}{\cm} and a thrust-to-weight ratio of 6.8. 
It is controlled by a low-level controller that takes \ac{CTBR} input.
The payload is carried by a 3D-printed release mechanism, shown in Fig.~\ref{fig:hardware}~(b)(c).
A loop is tied at the end of the rope that carries the payload. 
In the closed state, the shaft passes through the loop and retains the payload; on release, a servo retracts the shaft through a lever and linkage, driven by a binary signal from the flight controller.
The rope is \qty{33}{\cm} long, and the payload is a tennis ball with mocap markers, weighing \qty{85}{\gram}.

\vspace{-0.2cm}
\subsection{Quadrotor and Payload Hybrid Simulation}
\label{subsec:hybrid_sim}

\begin{figure*}[!t]
\centering
\usetikzlibrary{fit,backgrounds}
\begin{tikzpicture}[
  >={Latex[length=2mm,width=1.6mm]},
  font=\footnotesize,
  proc/.style={draw=blue!55, rounded corners=2pt, align=center, fill=blue!8,
               minimum height=8mm, text width=13.5mm, inner sep=2pt},
  field/.style={draw=teal!65, rounded corners=1.5pt, align=center, fill=teal!8,
                minimum height=8mm, text width=14mm, inner sep=2pt},
  intg/.style={draw=violet!55, rounded corners=2pt, align=center, fill=violet!10,
               fill opacity=0.55, draw opacity=1, text opacity=1,
               minimum height=5mm, inner sep=2pt, font=\scriptsize, text=violet!80},
  flow/.style={-{Latex[length=2mm]}, thick, blue!60},
  ord/.style={-{Latex[length=1.5mm]}, thin, gray!40},
  upd/.style={-{Latex[length=1.5mm]}, thin, teal!45},
  intgst/.style={{Latex[length=1.5mm]}-{Latex[length=1.5mm]}, thin, teal!45},
  rd/.style={-{Latex[length=1.5mm]}, thin, teal!45},
  cross/.style={-{Latex[length=2.6mm]}, very thick, black},
  loop/.style={-{Latex[length=2mm]}, thick, violet!70},
  rootband/.style={draw=orange!80, fill=orange!22, rounded corners=2pt, align=center,
                   font=\footnotesize\bfseries, text=orange!45!black,
                   minimum height=5mm, inner sep=1pt},
  rlink/.style={circle, draw=orange!80, fill=orange!35, minimum size=2.4mm, inner sep=0},
  pay/.style={circle, draw=orange!90, fill=orange!60, minimum size=5.2mm, inner sep=0},
  rchain/.style={orange!80, thick},
]
\node[proc] (cmd) {Command queue};
\node[proc, right=3.5mm of cmd] (llc)  {Low-level controller};
\node[proc, right=3.5mm of llc] (mot)  {Motor model};
\node[proc, right=3.5mm of mot] (bat)  {Battery model};
\node[proc, right=3.5mm of bat] (tt)   {Thrust--torque map};
\node[proc, right=3.5mm of tt]  (aero) {Aero\-dynamics};
\node[proc, right=3.5mm of aero](comp) {Composer};
\foreach \a/\b in {cmd/llc,llc/mot,mot/bat,bat/tt,tt/aero,aero/comp}
  \draw[ord] (\a) -- (\b);
\node[field, below=11mm of cmd] (kin) {Kinematics\\$\mathbf{p},\mathbf{q},\mathbf{v},\boldsymbol{\omega}$};
\node[field, right=3mm of kin]  (ill) {LLC integ.\\$\mathbf{i}_{\mathrm{llc}}$};
\node[field, right=3mm of ill]  (wde) {Motor des.\\$\boldsymbol{\omega}_{\mathrm{des}}$};
\node[field, right=3mm of wde]  (wmo) {Motor speed\\$\boldsymbol{\omega}_{\mathrm{mot}}$};
\node[field, right=3mm of wmo]  (uba) {Battery\\$u_{\mathrm{bat}}$};
\node[field, right=3mm of uba]  (wre) {Body wrench\\$\mathbf{F}_b,\boldsymbol{\tau}_b$};
\draw[intgst] (llc.south)  -- (ill.north);
\draw[intgst] (mot.south)  -- (wmo.north);
\draw[intgst] (bat.south)  -- (uba.north);
\draw[upd] ([xshift=2mm]llc.south)     -- ([xshift=-2mm]wde.north);
\draw[upd] ([xshift=-1.5mm]tt.south)   -- ([xshift=-3mm]wre.north);
\draw[upd] (aero.south)                -- (wre.north);
\draw[upd] ([xshift=1.5mm]comp.south)  -- ([xshift=3mm]wre.north);
\draw[rd] ([xshift=-2.5mm]kin.north) -- ([xshift=-2.5mm]llc.south);
\draw[rd] ([xshift=-1mm]uba.north)   -- ([xshift=2.5mm]llc.south);
\draw[rd] ([xshift=-2mm]wde.north)   -- ([xshift=-2mm]mot.south);
\draw[rd] ([xshift=-2mm]wmo.north)   -- ([xshift=-2mm]tt.south);
\draw[rd] ([xshift=1mm]wmo.north)    -- ([xshift=-2.5mm]aero.south);
\draw[rd] ([xshift=2.5mm]kin.north)  -- ([xshift=2.5mm]aero.south);
\node[intg, minimum width=72mm] (intg)
      at ($(ill.north)!0.5!(uba.north)+(0,5.5mm)$) {Integrator (Euler\,/\,RK4)};
\node[anchor=south west, inner sep=0] (drone3d) at ([xshift=17.6mm,yshift=7mm]wre.south east)
      {\includegraphics[width=28.9mm,trim=390 60 50 100,clip]{figures/kolibri_hd_w_wrench.png}};
\node[anchor=north west, inner sep=0] (wglyph)
      at ([xshift=-26mm,yshift=-0.5mm]drone3d.south) {\includegraphics[height=9mm]{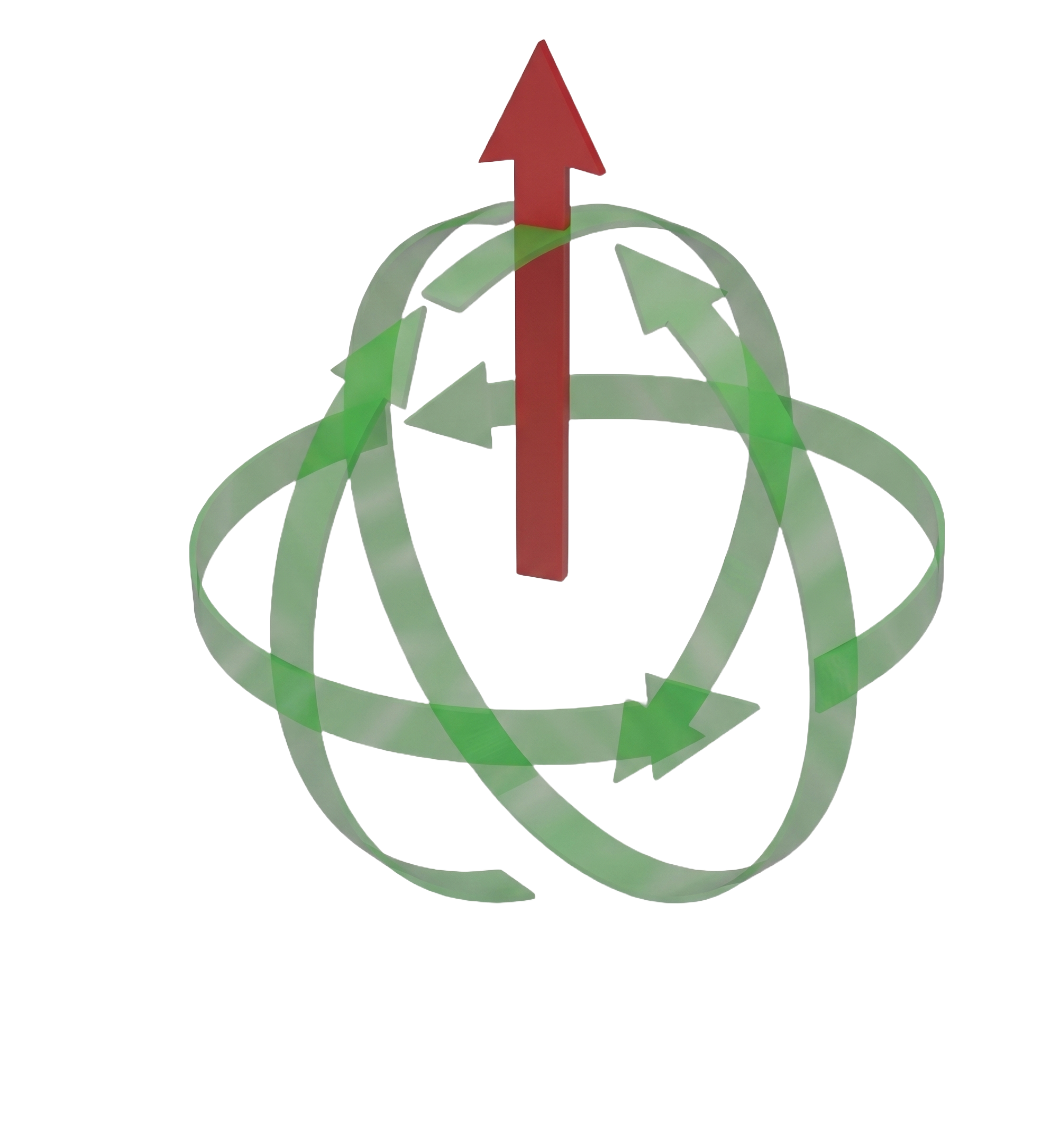}};
\node[anchor=west, text=blue!45!black]
      at ([xshift=-1.0mm]wglyph.east) {$=$ body wrench $(\mathbf{F}_b,\boldsymbol{\tau}_b)$};
\node[rotate=90, anchor=center, font=\scriptsize\bfseries, text=blue!60] (simlbl)
      at ([xshift=-4.5mm]cmd.west) {Sim. blocks};
\node[rotate=90, anchor=center, font=\scriptsize\bfseries, text=teal!60!black] (quadlbl)
      at ([xshift=-4.5mm]kin.west) {Quad state};
\begin{scope}[on background layer]
\node[draw=blue!40, rounded corners=3pt, fill=blue!3, inner sep=0.75mm,
      fit=(cmd)(comp)(kin)(wre)(drone3d)(simlbl)(quadlbl)] (ana) {};
\end{scope}
\node[anchor=south, font=\bfseries, text=blue!60]
      at ([yshift=-1mm]ana.north) {Standalone Quadrotor Simulator\, ($500\,$Hz)};
\node[draw=orange!75, rounded corners=3pt, fill=orange!8, align=center,
      minimum width=150mm, minimum height=21mm, below=7mm of ana, xshift=-4mm] (px) {};
\node[anchor=south, font=\bfseries, text=orange!60] at ([xshift=-16mm,yshift=-1mm]px.north)
      {PhysX articulation solver\,($500\,$Hz)};
\node[rootband, minimum width=114mm, anchor=north west] (proot)
      at ([xshift=4mm,yshift=-1.5mm]px.north west)
      {Quadrotor body \;\textemdash\; articulation floating root};
\node[draw=orange!80, fill=orange!45, minimum size=1.8mm, inner sep=0, anchor=north] (mnt)
      at ([xshift=10mm,yshift=-0.3mm]proot.south west) {};
\node[rlink] (l1) at ([xshift=13mm,yshift=-4.5mm]proot.south west) {};
\node[rlink] (l2) at ([xshift=18mm,yshift=-7mm]proot.south west) {};
\node[rlink] (l3) at ([xshift=23mm,yshift=-9mm]proot.south west) {};
\node[rlink] (l4) at ([xshift=28mm,yshift=-10.3mm]proot.south west) {};
\node[text=orange!70, font=\footnotesize] (ld) at ([xshift=33mm,yshift=-11mm]proot.south west) {$\cdots$};
\node[rlink] (l5) at ([xshift=38mm,yshift=-11.3mm]proot.south west) {};
\node[pay]   (pl) at ([xshift=44mm,yshift=-11.3mm]proot.south west) {};
\node[font=\scriptsize, text=orange!45!black, anchor=west] at ([xshift=1.4mm]pl.east) {Payload};
\draw[rchain] (mnt) -- (l1) -- (l2) -- (l3) -- (l4) -- (ld) -- (l5) -- (pl);
\node[font=\footnotesize\itshape, text=orange!45!black, align=left, anchor=north west]
      at ([xshift=65mm,yshift=-4mm]proot.south west)
      {$15$ passive universal joints\\
       (inextensible; slack in compression)};
\node[anchor=east, inner sep=0] (artic3d) at ([xshift=-10mm]px.east)
      {\includegraphics[height=18mm]{figures/simple_bodies_hd_w_wrench.png}};
\draw[cross] (wre.south) -- (wre.south |- proot.north);
\draw[cross] (kin.south |- proot.north) -- (kin.south);
\node[anchor=west, font=\footnotesize] at ($(wre.south |- px.north)+(1.5pt,3.5mm)$)
      {$\mathbf{F}_b,\,\boldsymbol{\tau}_b$};
\node[anchor=east, align=right, font=\footnotesize] at ($(kin.south |- px.north)+(-1.5pt,3.5mm)$)
      {Quadrotor \\ $\mathbf{p},\mathbf{q},\mathbf{v},\boldsymbol{\omega}$};

\node[above=6mm of cmd, align=center] (pol) {Policy action\\(CTBR, $50\,$Hz)};
\draw[flow, black] (pol.south) -- (ana.north -| cmd.north) -- (cmd.north);
\end{tikzpicture}
\caption{Hybrid simulation loop. The analytical quadrotor model maps delayed CTBR commands and current quadrotor state through the simulation modules to compute the body-frame wrench $(\mathbf{F}_b,\boldsymbol{\tau}_b)$, while integrating its internal states. The net wrench is applied to a PhysX reduced-coordinate articulation in which the quadrotor is the floating root of a rope-payload. PhysX advances the chain and writes the updated root pose and velocity back to the analytical model. The coupled loop runs at $500\,$Hz, holding each $50\,$Hz policy action for $10$ substeps, and exchanges only the body wrench and root kinematics.}
\label{fig:hybrid_loop}
\end{figure*}

The hybrid simulation, implemented in the IsaacLab framework~\cite{mittal2025isaaclab}, consists of two distinct parts, as shown in Fig.~\ref{fig:hybrid_loop}: a fast and accurate standalone quadrotor simulator implemented using Warp~\cite{Macklin_Warp_A_High-performance_2022}, and rope and payload dynamics simulated by a physics solver, in our case PhysX~\cite{nvidia_physx}.
\reb{The architecture preserves the identified response from \ac{CTBR} commands to body wrench while the solver integrates the coupled mechanics. Solver portability is an additional benefit (Sec.~\ref{subsec:sim_studies}).}

\paragraph{Standalone quadrotor model}
While the full suspended-payload system is hard to model analytically, the dynamics of the standalone quadrotor are relatively easy to identify. 
Based on the Agilicious \cite{foehn2022agilicious} architecture, the model contains a cascade of batched blocks.
As shown in Table~\ref{tab:analytical_blocks}, the blocks include an actuation-delay queue, a low-level controller, a motor model, a battery model, a map from motor speeds to rotor thrusts and drag torques, and an aerodynamic augmentation capturing drag and rotor inflow. 
\reb{All blocks are identified from real data of the standalone vehicle: the motor constants, thrust--torque map, and battery model from real-world measurements, and the aerodynamic effects from agile flight data.}
The blocks compose a net body wrench $(\mathbf{F}_b, \boldsymbol{\tau}_b)$ which can be directly integrated to simulate the standalone quadrotor. 
We provide an Euler and an RK4 integrator, allowing the standalone quadrotor simulation to be used for tasks that do not involve external forces, such as drone racing~\cite{song2023rl_vs_oc, xing2024multi}, navigation~\cite{Zhai2025PAMPPIPM}, obstacle avoidance~\cite{falanga2020obstacle}, and aerobatics~\cite{merk2026learningacrobaticflightpreferences, gao2025aerobatics}.
In this work, however, to simulate the suspended-payload system, we provide the resulting body-frame wrench to the physics solver.  
System identification is performed once on the standalone quadrotor and reused unchanged when the payload is attached.

\begin{table}[!t]
\centering
\caption{Blocks within the analytical quadrotor model.}
\label{tab:analytical_blocks}
\footnotesize
\begin{tabularx}{\columnwidth}{lX}
\toprule
\textbf{Block} & \textbf{Function / Output} \\
\midrule
\grayrow
Command queue        & Buffers \ac{CTBR} command to account for transmission and actuation delay $\tau_d$ \\
Low-level controller & Body-rate control and voltage-compensated thrust mapping $\to$ motor-speed setpoints \\ \grayrow
Motor model          & First-order model with time constant $\tau_{\text{mot}}$ \\
Battery model       & Pack voltage $u_{\mathrm{bat}}$ from remaining charge and drawn power (load-dependent sag)\\ \grayrow
Thrust-torque map   & Motor speed $\to$ rotor thrust $f_i$ and drag torque $\tau_i$\\ 
Aerodynamics         & Drag and rotor inflow effects $\Delta\mathbf{F},\Delta\boldsymbol{\tau}$ \\ \grayrow
Composer             & Net wrench $(\mathbf{F}_b, \boldsymbol{\tau}_b)$ to integrator \\ 
\bottomrule
\end{tabularx}
\vspace{-0.3cm}
\end{table}

\paragraph{Articulated payload and coupling}
The rope and payload are authored as a chain of $15$ rigid segments terminating in a rigid payload link, attached to the quadrotor via a fixed mount, similar to~\cite{zeng2025decentralizedaerialmanipulationcablesuspended}.
The $2$-DOF universal joints between rope segments are passive and effectively frictionless, so the chain behaves as a discretized inextensible tether under tension and goes slack in compression.
The PhysX reduced-coordinate articulation solver integrates this chain together with the quadrotor body as its floating root.
At every simulation step, the analytical quadrotor model first writes $(\mathbf{F}_b, \boldsymbol{\tau}_b)$ onto the quadrotor link, and the solver then advances the full articulation under that wrench, gravity, and the joint-constraint reactions. 
The reaction wrench at the rope mount is transmitted back into the quadrotor body automatically by the solver, with no explicit coupling term, and the quadrotor kinematic state for the next step is read directly from the solver, closing the loop (Fig.~\ref{fig:hybrid_loop})\reb{; no second quadrotor trajectory is integrated}.
This coupled exchange runs at \qty{500}{\hertz}: each \qty{50}{\hertz} policy action is held fixed while the analytical model and the articulation solver advance 10 substeps before the next action is queried.

\paragraph{Release event and post-release dynamics}
The policy outputs a release scalar at each control step; when it exceeds a threshold, the payload is released after a delay matching the identified actuation lag ($\qty{0.11}{\second}$ for our mechanism). 
The payload state at release is used to predict projectile motion analytically using a closed-form parabolic trajectory. 
The episode is then terminated, and the predicted landing location is used to calculate the reward.

\subsection{Reinforcement Learning Training}
\label{subsec:rl_training}
We train the throwing policy with PPO~\cite{schulman2017proximal} inside the hybrid simulator, with an asymmetric actor-critic setup. At the start of each episode, the target landing location $\mathbf{x}_g$ is sampled uniformly on a ground-plane disk of radius $\qty{2}{\meter}$ centered on the spawn point.

\paragraph{Action and observation spaces}
At each control step ($\qty{50}{\hertz}$), the actor outputs a 5-dimensional command
$
\mathbf{u}_t = (c_t,\,\boldsymbol{\omega}^{\mathrm{cmd}}_t,\,\rho_t) \in \mathbb{R}^5,
$
where $c_t$ is the collective-thrust setpoint, $\boldsymbol{\omega}^{\mathrm{cmd}}_t \in \mathbb{R}^3$ is the body-rate setpoint, and $\rho_t$ is a release latch that trips when $\rho_t > 0.95$. \reb{The actor observation $o_t \in \mathbb{R}^{35}$ comprises the quadrotor position~($3$), linear velocity~($3$), and angular velocity~($3$), all in the world frame; attitude~($6$) in a 6D rotation representation~\cite{zhou2019continuity}; the previous \ac{CTBR} command $(c_{t-1},\boldsymbol{\omega}^{\mathrm{cmd}}_{t-1})$ ($4$); the goal-to-payload offset ($3$); the payload-to-quadrotor relative position \rev{$\Delta\mathbf{p}_t$} ($3$) \rev{and its values at the three preceding control steps, $(\Delta\mathbf{p}_{t-3},\Delta\mathbf{p}_{t-2},\Delta\mathbf{p}_{t-1})$} ($9$); and an attached/released indicator~($1$). The critic additionally receives the payload world-frame velocity $\mathbf{v}_p$ as privileged observation.}

\paragraph{Reward}
The instantaneous reward decomposes as
$
r_t = r_t^{\mathrm{carry}} + r_t^{\mathrm{rel}} + r_t^{\mathrm{neg}}.
$
Let $\mathbf{1}_{\mathrm{att}}$, $\mathbf{1}_{\mathrm{rel}}$, and $\mathbf{1}_{\mathrm{neg}}$ denote the payload-attached, release-step, and negative-termination indicators; $\mathbf{v}_q, \boldsymbol{\omega}_q$ are the quadrotor linear and angular velocities and $\mathbf{v}_p$ the payload velocity in world-frame; and $\Delta u_j = u_t^j - u_{t-1}^j$ is the per-step change of the $j$-th component of $\mathbf{u}_t$, i.e. $u^1_t = c_t$ and $ u^{2:4}_t = \boldsymbol{\omega}^{\mathrm{cmd}}_t$.

The carry reward, active while the payload is attached, further decomposes into a kinetic term, a command-magnitude term, a command-rate term, and a carry-time term:
\begin{equation}
\begin{aligned}
r_t^{\mathrm{carry}} &= r_t^{\mathrm{kin}} + r_t^{\mathrm{cmd}} + r_t^{\mathrm{rate}} + r_t^{\mathrm{time}}, \\
r_t^{\mathrm{kin}}   &= \mathbf{1}_{\mathrm{att}}\!\left[\alpha_v\|\mathbf{v}_q\|^2 + \alpha_\omega\|\boldsymbol{\omega}_q\|^2 + \alpha_p\|\mathbf{v}_p - \mathbf{v}_q\|^2\right], \\
r_t^{\mathrm{cmd}}   &= -\mathbf{1}_{\mathrm{att}}\,\beta \sum_{i=x,y,z} c^{\omega}_i\, (\omega^{\mathrm{cmd}}_i)^2, \\
r_t^{\mathrm{rate}}  &= -\mathbf{1}_{\mathrm{att}}\,\beta \sum_{j=1}^{4} c^{\Delta}_j\!\left(\eta\,|\Delta u_j| + (\Delta u_j)^2\right), \\
r_t^{\mathrm{time}}  &= -\gamma_t\, \Delta t_{\mathrm{ctrl}}\, \mathbf{1}_{\mathrm{att}\,\lor\,\mathrm{rel}}.
\end{aligned}
\label{eq:r_carry}
\end{equation}
$r^{\mathrm{kin}}$ penalizes high quadrotor body speeds and large relative payload motion via the relative-velocity term. $r^{\mathrm{cmd}}$ discourages large body-rate commands. $r^{\mathrm{rate}}$ damps step-to-step command changes using a mixed $L_1{+}L_2$ form: $L_2$ alone leaves persistent low-amplitude jitter, while $L_1$ alone induces deadbanding. $r^{\mathrm{time}}$ is a constant per-step cost incurred while carrying or at the release step, pushing the policy to commit to a throw rather than hover indefinitely. All coefficients are listed in Table~\ref{tab:reward_coeffs}.

The task signal is concentrated in the release reward, active only at release. It decomposes into a landing-accuracy term, a time-of-flight term, and a proximity-guard term:
\begin{equation}
\begin{aligned}
r_t^{\mathrm{rel}}  &= r_t^{\mathrm{land}} + r_t^{\mathrm{tof}} + r_t^{\mathrm{prox}}, \\
r_t^{\mathrm{land}} &= \mathbf{1}_{\mathrm{rel}}\, k_\ell\, e^{-\lambda \|\hat{\mathbf{x}}_{\mathrm{imp}} - \mathbf{x}_g\|}, \\
r_t^{\mathrm{tof}}  &= -\mathbf{1}_{\mathrm{rel}}\, k_h\, t_{\mathrm{hit}}, \\
r_t^{\mathrm{prox}} &= -\mathbf{1}_{\mathrm{rel}}\, k_p\, g_{q}.
\end{aligned}
\label{eq:r_rel}
\end{equation}
Here $\mathbf{x}_g$ is the goal location, $\hat{\mathbf{x}}_{\mathrm{imp}}$ is the predicted impact point, and $t_{\mathrm{hit}}$ is the predicted time to impact; both are obtained in closed form from the release-instant payload position and velocity under projectile motion. $r^{\mathrm{land}}$ rewards proximity of the predicted impact to the goal. $r^{\mathrm{tof}}$ penalizes long parabolic trajectories, encouraging more energetic throws. $r^{\mathrm{prox}}$ uses the guard $g_q$ to penalize post-release payload trajectories that pass too close to the quadrotor, preventing collision:
\begin{equation}
g_{q} = \max\!\left( 0,\; d_m - \!\!\min_{s \in [0, t_{\mathrm{hit}}]}\! \|\mathbf{q}_p(s) - \mathbf{p}_q^{r}\| \right),
\label{eq:g_q}
\end{equation}
where $\mathbf{q}_p(s)$ is the analytical post-release payload trajectory sampled at $K=20$ fractions of $t_{\mathrm{hit}}$, $\mathbf{p}_q^{r}$ is the quadrotor position at release, and $d_m$ is a minimum-clearance margin.

Negative termination incurs a constant penalty,
\begin{equation}
r_t^{\mathrm{neg}} = -k_d\, \mathbf{1}_{\mathrm{neg}}.
\label{eq:r_neg}
\end{equation}
An episode terminates positively on a successful release, and negatively (firing $\mathbf{1}_{\mathrm{neg}}$) on the quadrotor leaving altitude or horizontal bounds, on excessive tilt, on a payload-quadrotor collision, on the payload rising above the quadrotor while still attached, or at a $\qty{5}{\second}$ timeout.

By construction, $r_t^{\mathrm{rel}}$ is sparse in time but dense in the impact error, and coupling the target-accuracy bonus with the time-to-impact penalty removes the pathology in which the policy prematurely releases to avoid paying per-step costs.

\paragraph{Domain randomization}
At every episode reset during training, the quadrotor body mass is perturbed by $\pm5\%$ and the platform is initialized uniformly over position ($\pm0.5/0.5/0.3\,\si{\meter}$), linear velocity ($\pm0.5/0.5/0.3\,\si{\meter\per\second}$), attitude ($\pm20^{\circ}$ per axis), and body rate ($\pm30^{\circ}\!/\si{\second}$ per axis). Variation in the suspended-payload dynamics is introduced in two ways: (i) each parallel environment samples one of six tether-length variants spanning $\pm6\%$ about the nominal cable; and (ii) at every reset, the cable is articulated from the quadrotor through a vertical-axis (yaw) joint angle that selects the azimuth of cable tilt followed by a horizontal-axis (roll) joint angle that sets the tilt magnitude, both joints perturbed by uniform angle offsets of $\pm15^{\circ}$ and initial rates of $\pm15^{\circ}\!/\si{\second}$. Sensor and actuator imperfections are modeled as additive Gaussian noise resampled every step and bias sampled once per episode, applied to the unnormalized observations in SI units (std $0.01$ and $0.001$) and to the raw policy outputs before their mapping to commands (std $0.025$ and $0.02$). The payload mass, aerodynamic drag, thrust map, and actuator latency are deliberately fixed to values measured on the real platform, isolating the randomization to quantities for which the sim-to-real mismatch is known to dominate.

\paragraph{Training setup}
\reb{PPO is used to train the policy on a single RTX5090 GPU with $4096$ parallel environments. A full run of $1500$ update iterations ($\approx 2\!\times\!10^{8}$ environment steps) takes about \qty{70}{\minute}. The network architecture and PPO hyperparameters are listed in Table~\ref{tab:ppo_params}.}

\begin{table}[!t]
\centering
\caption{Reward coefficients.}
\label{tab:reward_coeffs}
\footnotesize
\setlength{\tabcolsep}{2pt}
\begin{tabular}{@{}lll@{}}
\xtoprule{3}
\textbf{Symbol} & \textbf{Value} & \textbf{Meaning} \\
\xmidrule{3}
\grayrow
$\alpha_v$            & $-10^{-3}$                       & Quadrotor linear-velocity weight \\
$\alpha_\omega$       & $-10^{-3}$                       & Quadrotor angular-velocity weight \\ \grayrow
$\alpha_p$            & $-3{\times}10^{-3}$              & Payload-quad relative-velocity weight \\
$\beta$               & $0.5$                            & Smoothness penalty scale \\ \grayrow
$c^{\omega}$          & $(0.05,\,0.05,\,0.05)$           & Body-rate command weights \\
$c^{\Delta}$          & $(0.15,\,0.025,\,0.025,\,0.025)$ & Command-rate weights \\ \grayrow
$\eta$                & $0.2$                            & $L_1{/}L_2$ mix ratio \\
$\gamma_t$            & $\qty{2.0}{\per\second}$          & Per-second carry-time cost \\ \grayrow
$z_{\mathrm{land}}$   & $\qty{0.04}{\meter}$              & Landing plane height \\
$d_m$                 & $\qty{0.25}{\meter}$              & Quadrotor-proximity margin \\ \grayrow
$K$                   & $20$                             & Post-release trajectory samples \\
$k_\ell$              & $500$                            & Landing-accuracy bonus \\ \grayrow
$\lambda$             & $\qty{2.5}{\per\meter}$           & Accuracy-decay rate \\
$k_h$                 & $2.0$                            & Time-of-flight penalty \\ \grayrow
$k_p$                 & $100$                            & Proximity penalty \\
$k_d$                 & $50$                             & Negative-termination penalty \\
\xbottomrule{3}
\end{tabular}
\end{table}

\begin{table}[!t]
\centering
\caption{\reb{Network architecture and PPO hyperparameters.}}
\label{tab:ppo_params}
\footnotesize
\setlength{\tabcolsep}{0pt}
\begin{tabular}{@{}p{0.256\columnwidth}p{0.296\columnwidth}p{0.314\columnwidth}p{0.134\columnwidth}@{}}
\xtoprule{4}
\reb{\textbf{Parameter}} & \reb{\textbf{Value}} & \reb{\textbf{Parameter}} & \reb{\textbf{Value}} \\
\xmidrule{4}
\grayrow
\reb{Actor/critic net} & \reb{MLP $3{\times}256$, ELU} & \reb{Discount $\gamma$} & \reb{$0.99$} \\
\reb{Environments} & \reb{$4096$} & \reb{GAE $\lambda$} & \reb{$0.95$} \\
\grayrow
\reb{Rollout steps} & \reb{$32$} & \reb{Clip ratio $\epsilon$} & \reb{$0.2$} \\
\reb{Epochs} & \reb{$5$} & \reb{Value-loss weight} & \reb{$1.0$} \\
\grayrow
\reb{Mini-batches} & \reb{$4$} & \reb{Entropy weight} & \reb{$0.0025$} \\
\reb{Learning rate} & \reb{$5{\times}10^{-4}$, adaptive} & \reb{Max.\ gradient norm} & \reb{$1.0$} \\
\grayrow
\reb{Target KL} & \reb{$0.016$} & \reb{Initial action std} & \reb{$0.5$} \\
\xbottomrule{4}
\end{tabular}
\vspace{-0.25cm}
\end{table}

%% file: sections/experiments.tex
\section{Experiments}
\begin{figure*}[!t]
\centering
\setlength{\tabcolsep}{2pt}
\renewcommand{\arraystretch}{1.0}
\tikzset{ts/.style={font=\scriptsize, fill=white, fill opacity=0.7, text opacity=1, inner sep=1.2pt, rounded corners=1pt, anchor=south}}
\newcommand{\throwpanel}[3]{
\begin{tikzpicture}
\node[anchor=south west,inner sep=0] (img) {\includegraphics[#1]{#2}};
\begin{scope}[x={(img.south east)},y={(img.north west)}]#3\end{scope}
\end{tikzpicture}}
\begin{tabular}{@{}c@{\hspace{0.025\textwidth}}c@{}}
\begin{tabular}[c]{@{}c@{\hspace{0.03\textwidth}}c@{}}
\throwpanel{width=0.32\textwidth, trim={460 310 480 40}, clip}{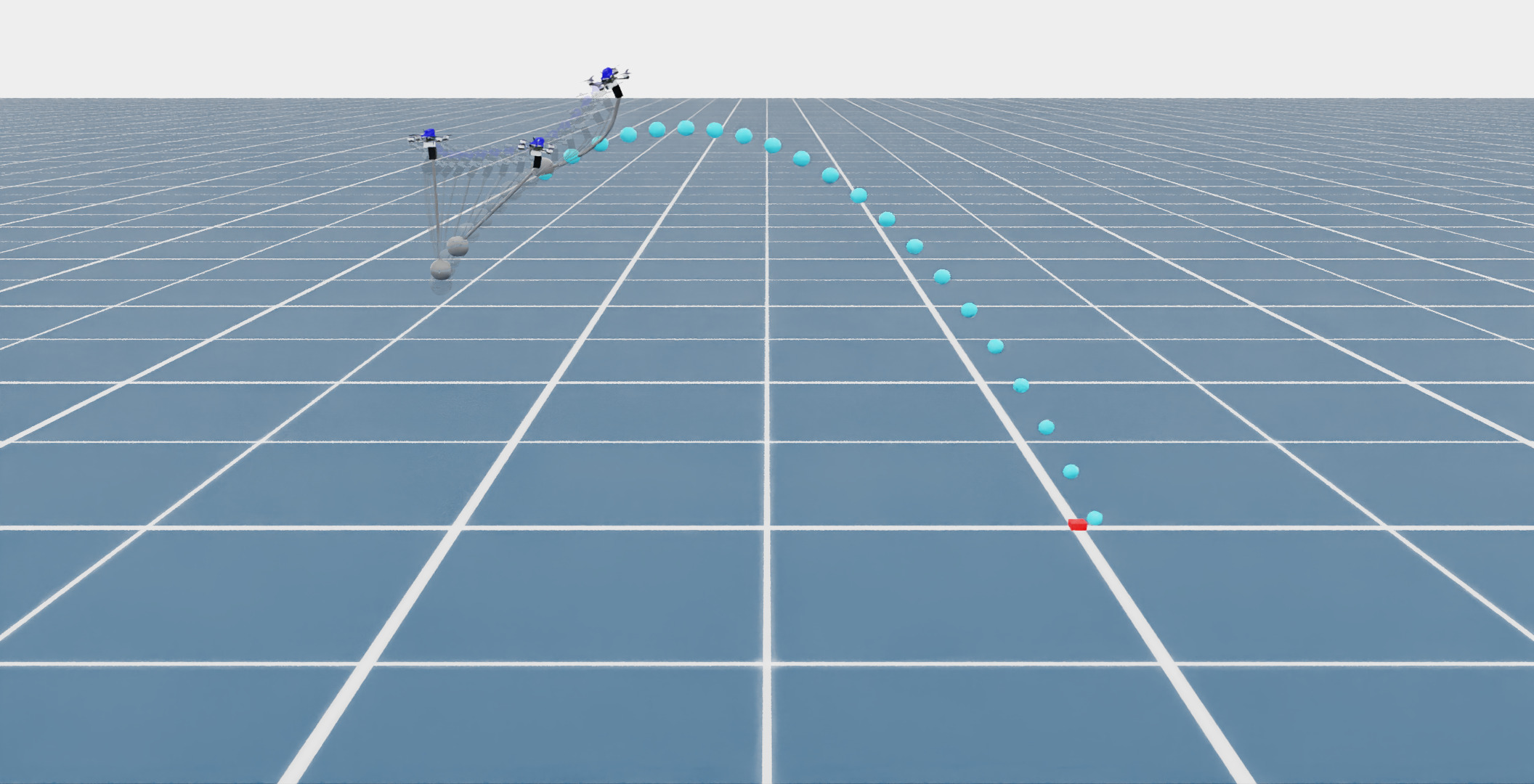}{%
  \node[ts] at (0.110,0.828) {$0\,$s};
  \node[ts] at (0.226,0.822) {$0.15\,$s};
  \node[ts] at (0.415,0.865) {$0.4\,$s};
  \node[ts,anchor=south east] at (0.835,0.035) {$t_{\mathrm{land}}\!\approx\!1.1\,$s};} &
\throwpanel{width=0.32\textwidth, trim={380 70 340 260}, clip}{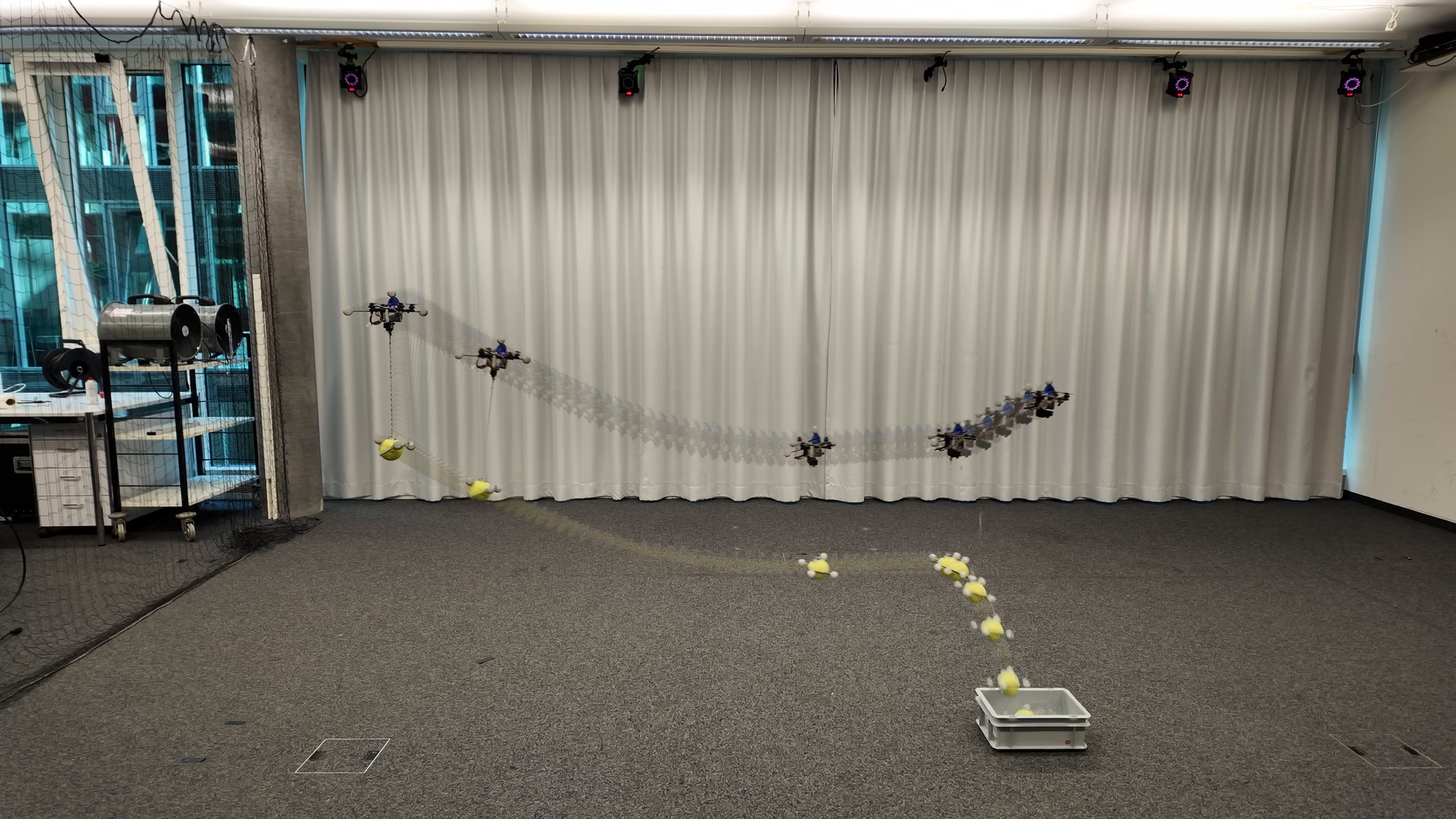}{%
  \node[ts] at (0.114,0.865) {$0\,$s};
  \node[ts] at (0.229,0.788) {$1\,$s};
  \node[ts] at (0.578,0.635) {$2\,$s};
  \node[ts] at (0.733,0.649) {$2.5\,$s};
  \node[ts,anchor=south east] at (0.735,0.04) {$t_{\mathrm{land}}\!\approx\!2.8\,$s};} \\[1pt]
{\small (a) Ours, simulation} & {\small (b) AD-Planner~\cite{AD_planner}, real world} \\[3pt]
\throwpanel{width=0.32\textwidth, trim={430 220 500 150}, clip}{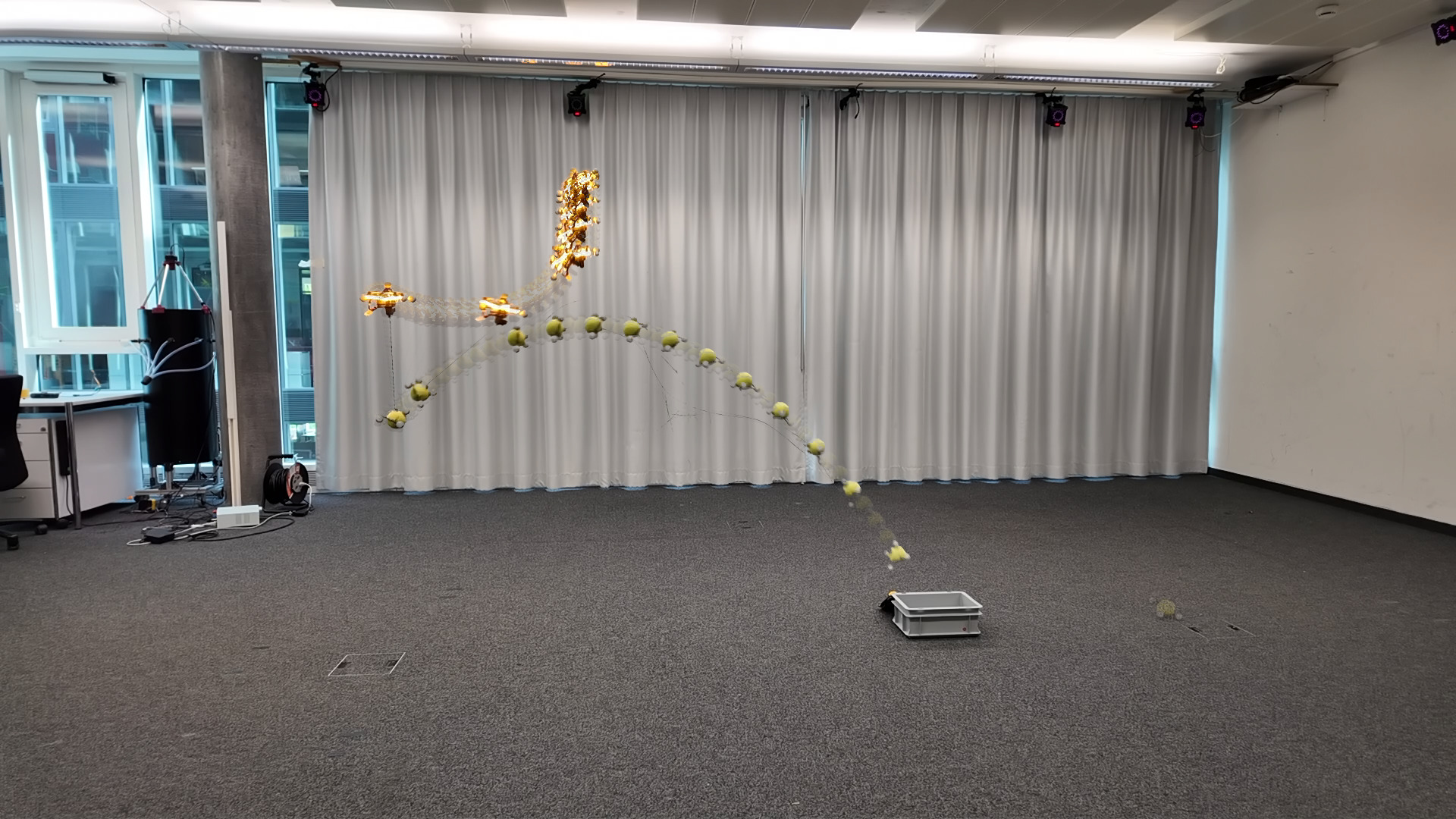}{%
  \node[ts] at (0.078,0.717) {$0\,$s};
  \node[ts] at (0.230,0.687) {$0.15\,$s};
  \node[ts] at (0.419,0.705) {$0.4\,$s};
  \node[ts,anchor=south east] at (0.730,0.040) {$t_{\mathrm{land}}\!\approx\!1.1\,$s};} &
\throwpanel{width=0.32\textwidth, trim={350 100 380 120}, clip}{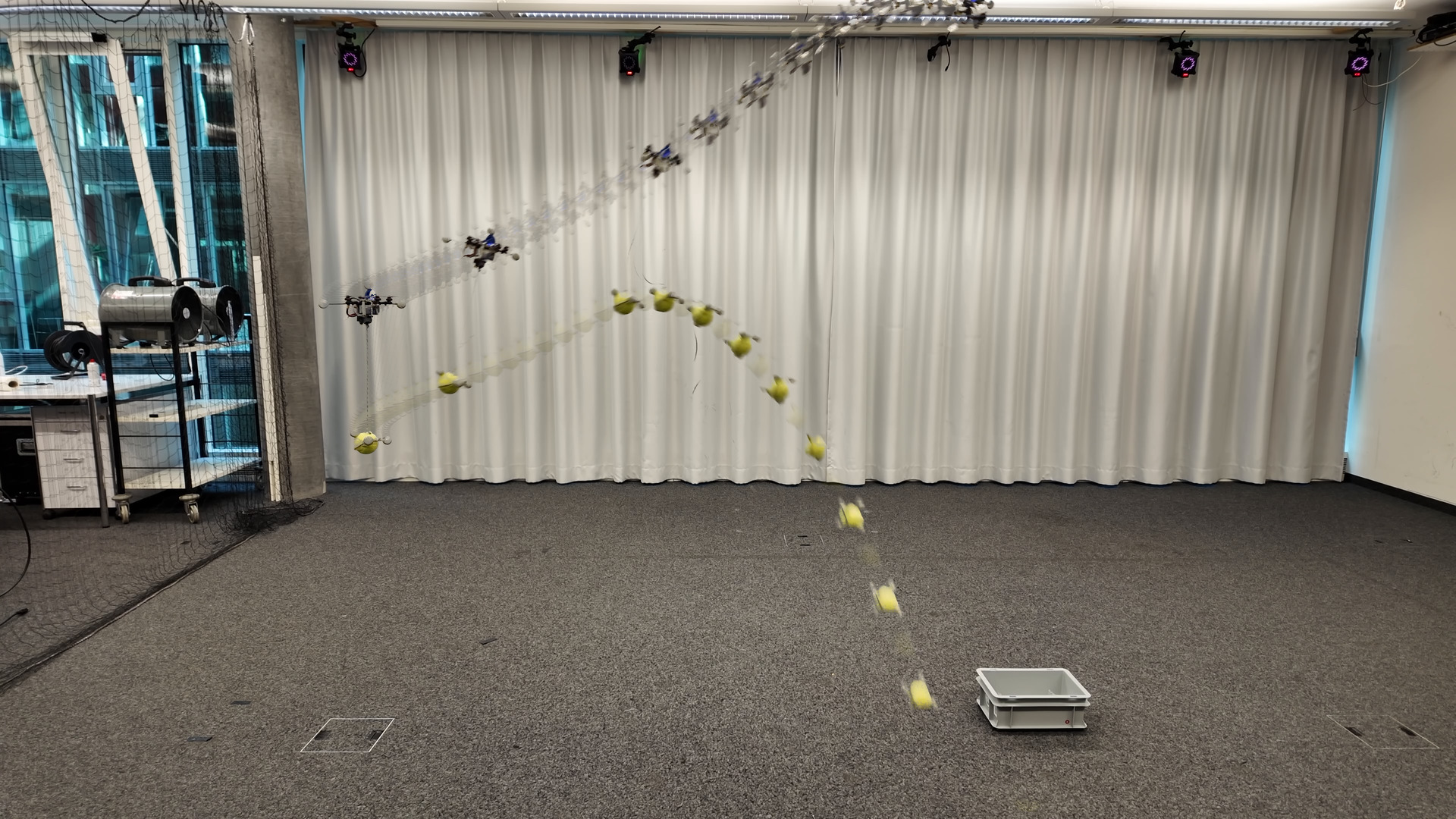}{%
  \node[ts] at (0.113,0.723) {$0\,$s};
  \node[ts] at (0.233,0.812) {$0.8\,$s};
  \node[ts] at (0.438,0.925) {$1.2\,$s};
  \node[ts,anchor=south east] at (0.680,0.030) {$t_{\mathrm{land}}\!\approx\!1.7\,$s};} \\[1pt]

{\small (c) Ours, real world} & {\small (d) Foehn~\cite{Foehn_Falanga_Kuppuswamy_Tedrake_Scaramuzza_2017}, real world} \\
\end{tabular} &
\begin{tabular}[c]{@{}c@{}}
\includegraphics[width=0.31\textwidth, trim={8 7 7 7}, clip]{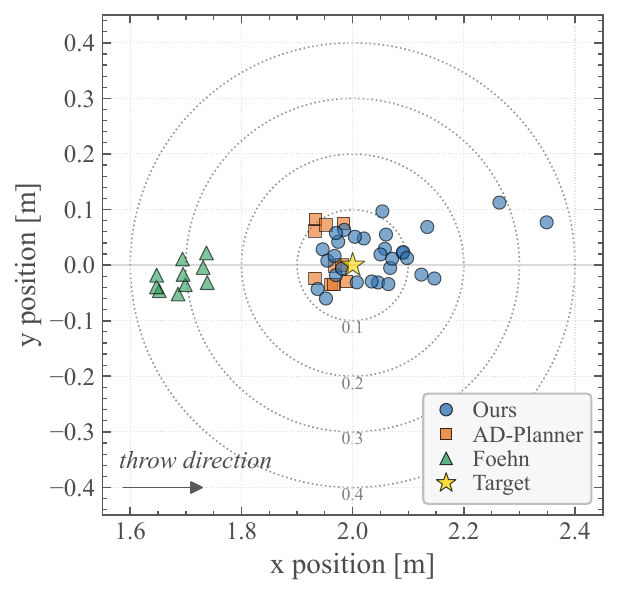} \\[1pt]
{\small (e) Landing scatter at the \qty{2.0}{\meter} target} \\
\end{tabular}
\end{tabular}
\caption{\reb{Representative throws to the \qty{2.0}{\meter} target. Timestamps are relative to policy onset; the last timestamp beside the quadrotor marks release, and $t_{\mathrm{land}}$ marks the landing of the payload. (a), (c): our policy in simulation and on hardware. (b), (d): the model-based baselines executing the same task; AD-Planner~(b) delivers accurately via a \qty{2.8}{\second} conservative cast, while Foehn's trajectory optimization (d) releases \qty{1.2}{\second} after the throw starts and undershoots the target. (e): top-down landing spots of all hardware throws to this target, thirty for our policy and ten per baseline.}}
\label{fig:throw_sequence}
\end{figure*}



\subsection[Simulation Studies]{\reb{Simulation Studies}}
\label{subsec:sim_studies}
\reb{Each evaluation in simulation includes 8192 throws with all training randomization of Sec.~\ref{subsec:rl_training} active.}
We train five policies with different random seeds.
The mean landing errors lie between \qty{4.34}{\centi\meter} and \qty{6.99}{\centi\meter}.
The best (\qty[uncertainty-mode=separate]{4.34 \pm 2.69}{\centi\meter}) is used in the robustness study below and deployed on hardware.

\begin{figure}[!t]
\centering
\subfloat[\reb{Payload mass [\%]}\label{fig:sim_mass}]{\includegraphics{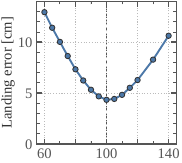}}\hspace{4pt}%
\subfloat[\reb{Std.\ [$\times$ training]}\label{fig:sim_noise}]{\includegraphics{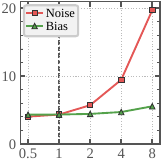}}\hspace{4pt}%
\subfloat[\reb{Target distance [m]}\label{fig:sim_distance}]{\includegraphics{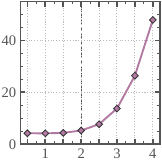}}
\caption{\reb{Mean landing error in simulation as a function of (a) the payload mass, (b) the standard deviation of the observation noise and bias, relative to its training value, and (c) the target distance.}}
\label{fig:sim_robustness}
\end{figure}

\reb{\textit{Robustness and limits.}
We test the deployed policy under conditions that deviate from training in three ways (Fig.~\ref{fig:sim_robustness}).
First, the payload mass, which is fixed during training, is varied from its nominal value, and the landing error grows only gradually (Fig.~\ref{fig:sim_mass}), demonstrating robustness to small mass variations.
Second, we scale the observation noise and bias of Sec.~\ref{subsec:rl_training}; per-step noise matters more in our setup, since a constant bias can be calibrated out, so we train with a bias ten times smaller than the noise.
Even at $4\times$ the training noise, corresponding to a standard deviation of \qty{4}{\centi\meter} on every position observation, the policy maintains a mean landing error below \qty{10}{\centi\meter} (Fig.~\ref{fig:sim_noise}).
Finally, we move the target beyond the \qty{2}{\meter} training radius: the policy extrapolates to the out-of-distribution \qty{2.5}{\meter} target with a moderately larger landing error, but degrades quickly for farther targets (Fig.~\ref{fig:sim_distance}).}

\reb{\textit{Solver independence.}
To verify that our hybrid simulator is not tied to a specific physics solver, we replace PhysX with Newton~\cite{newton2025engine} and retrain a policy with all other settings unchanged.
We then evaluate the deployed PhysX-trained policy and the Newton-trained policy in both solvers.
As shown in Table~\ref{tab:solver_xeval}, evaluating a policy in the solver it was not trained in increases its mean landing error by less than \qty{0.3}{\centi\meter}, and the two policies reach similar accuracy in the solvers they were trained in.}

\begin{table}[!t]
\centering
\caption{\reb{Landing error [\si{\centi\meter}] in simulation of policies trained in PhysX and in Newton, each evaluated in both solvers over 8192 throws.}}
\label{tab:solver_xeval}
\footnotesize
\setlength{\tabcolsep}{6pt}
\begin{tabular}{@{}lcc@{}}
\xtoprule{3}
\textbf{Trained in} & \textbf{Evaluated in PhysX} & \textbf{Evaluated in Newton} \\
\xmidrule{3}
\grayrow
PhysX (deployed) & $4.34 \pm 2.69$ & $4.60 \pm 2.85$ \\ 
Newton & $4.59 \pm 2.85$ & $4.47 \pm 2.77$ \\
\xbottomrule{3}
\end{tabular}
\end{table}

\subsection{Real-World Experiments}
\label{subsec:throwing_experiments}
To ensure repeatability of the experiments on hardware, the throwing is initiated from hover, with the payload around $z=\qty{0.95}{\meter}$ at the origin.
The policy is commanded to throw forward in the $x$ direction, at targets \qty{1.5}{\meter}, \qty{2.0}{\meter}, and \qty{2.5}{\meter} away from the origin. Note that the last target is out-of-distribution and was not seen during training. 
We use two metrics for the evaluation: the time between policy activation and payload landing to evaluate agility and efficiency, and the landing-point distance from the goal to measure accuracy.

\reb{\textit{Baselines.}
We reproduce the complementarity-constrained trajectory optimization of Foehn~\textit{et al.}~\cite{Foehn_Falanga_Kuppuswamy_Tedrake_Scaramuzza_2017} and the error-bounded delivery planner AD-Planner~\cite{AD_planner} using their default parameters.
We retain Foehn's complementarity formulation and AD-Planner's error-bounded release constraints, adapting platform parameters and task inputs to our setup. We verify the implementations against their published scenarios. Both planners use the same tracking controller, following Foehn's implementation. All methods use the same quadrotor, cable, payload, release mechanism, motion-capture state estimate, hover start, and targets. Duration is measured from policy or trajectory execution to landing, excluding planning time.}

\begin{table}[!t]
\centering
\caption{\reb{Throwing accuracy and agility on hardware over thirty throws per target for our policy and ten per target for each baseline. }}
\label{tab:throwing_results}
\footnotesize
\setlength{\tabcolsep}{5pt}
\begin{tabular}{@{}llcc@{}}
\xtoprule{4}
\reb{\textbf{Target}} & \reb{\textbf{Method}} &
\reb{\textbf{Landing Error [\si{\meter}]}} &
\reb{\textbf{Throw Duration [\si{\second}]}} \\
\xmidrule{4}
\reb{\multirow{3}{*}{\qty{1.5}{\meter}}} & \cellcolor{gray!10}\reb{Ours (RL)}    & \cellcolor{gray!10}\reb{$0.079 \pm 0.030$} & \cellcolor{gray!10}\reb{$1.084 \pm 0.013$} \\
 & \reb{Foehn}        & \reb{$0.252 \pm 0.056$}          & \reb{$1.627 \pm 0.007$} \\
 & \reb{AD-Planner}   & \reb{$0.074 \pm 0.018$} & \reb{$2.399 \pm 0.041$} \\
\xmidrule{4}
\reb{\multirow{3}{*}{\qty{2.0}{\meter}}} & \cellcolor{gray!10}\reb{Ours (RL)}    & \cellcolor{gray!10}\reb{$0.088 \pm 0.070$} & \cellcolor{gray!10}\reb{$1.142 \pm 0.014$} \\
 & \reb{Foehn}        & \reb{$0.309 \pm 0.034$}          & \reb{$1.773 \pm 0.011$} \\
 & \reb{AD-Planner}   & \reb{$0.062 \pm 0.028$} & \reb{$2.737 \pm 0.040$} \\
\xmidrule{4}
\reb{\multirow{3}{*}{\qty{2.5}{\meter}}} & \cellcolor{gray!10}\reb{Ours (RL)}    & \cellcolor{gray!10}\reb{$0.112 \pm 0.069$} & \cellcolor{gray!10}\reb{$1.168 \pm 0.020$} \\
 & \reb{Foehn}        & \reb{$0.424 \pm 0.041$}          & \reb{$2.189 \pm 0.012$} \\
 & \reb{AD-Planner}   & \reb{$0.053 \pm 0.043$} & \reb{$2.910 \pm 0.076$} \\
\xbottomrule{4}
\end{tabular}
\end{table}

\reb{\textit{Sim-to-real consistency.}
Compared with the simulated errors, the hardware landing errors of our policy are about \qty{3.6}{\centi\meter} larger at all three targets, including the out-of-distribution one.
Their spread is dominated by a few long throws (Fig.~\ref{fig:throw_sequence}(e)).}

\reb{\textit{Accuracy--agility trade-off.}
In Table~\ref{tab:throwing_results}, the model-based baselines sit at opposite ends of an accuracy--agility trade-off, which follows from how each treats execution error.
Foehn's method executes an aggressive nominal plan with a controller that regulates only the quadrotor, so nothing corrects the payload's state before release: it is the more agile of the two but the least accurate, undershooting increasingly with distance (Fig.~\ref{fig:throw_sequence}(e)).
AD-Planner bounds the effect of execution error on the landing point, which restricts its release states: it is the most accurate but the slowest, releasing only after approaching the target (Fig.~\ref{fig:throw_sequence}(b)).
Our policy is not bound to this choice: trained on the full dynamics, it commands the vehicle in closed loop from the observed quadrotor and payload states, without a precomputed release state or conservative constraints.
It is the fastest at every target, \rev{with an accuracy within \qty{3}{\centi\meter} of AD-Planner's inside the training radius}, advancing the accuracy--agility frontier.}

\reb{\textit{Analysis of the baselines.}
The measured release states explain the baselines' behaviors. For Foehn's method, the release is on time and the quadrotor tracks its reference closely, yet the payload leaves at only $76$--$83\%$ of the planned speed.
Ballistic propagation of the measured release state reproduces the landing points to within millimeters; the speed deficit therefore accounts for the undershoot and for its growth with distance.
A comparable undershoot is reported in~\cite{Foehn_Falanga_Kuppuswamy_Tedrake_Scaramuzza_2017}.
AD-Planner's release state is constrained by a certified landing-error bound (\qty{12}{\centi\meter} as published) that grows with the free-fall time and the release speed.
This bound is active at every target, whereas none of its velocity or acceleration limits is: the planner casts the payload from \qty{0.26}{\meter} away at a height of \qtyrange{0.5}{0.7}{\meter}, a free fall of less than \qty{0.3}{\second}.
Its duration is thus set by its constraints rather than by the platform; under the published limits, the certifiable casting range is below \qty{0.6}{\meter}, so a throw over \qtyrange{1.5}{2.5}{\meter} is infeasible.}

\begin{table}[!t]
\centering
\caption{\reb{Peak quantities during hardware throws to the \qty{2.5}{\meter} target, averaged over episodes. The specific thrust is the total rotor thrust divided by the vehicle mass. The last row gives the same peaks over a racing flight with a standalone quadrotor of the same model~\cite{geles2026racing}.}}
\label{tab:agility_metrics}
\scriptsize
\setlength{\tabcolsep}{5pt}
\renewcommand{\arraystretch}{1.05}
\begin{tabular}{@{}lccc@{}}
\xtoprule{4}
\reb{\textbf{Method}} &
\reb{\shortstack{\textbf{Specific thrust}\\$f_T$ $[\si{\meter\per\second\squared}]$}} &
\reb{\shortstack{\textbf{Body rate}\\$\lVert\boldsymbol{\omega}\rVert$ $[\si{\radian\per\second}]$}} &
\reb{\shortstack{\textbf{Payload $v$ at release}\\$\lVert\mathbf{v}_p\rVert$ $[\si{\meter\per\second}]$}} \\
\xmidrule{4}
\grayrow
\reb{Ours}       & \reb{$39.1 \pm 2.7$} & \reb{$10.5 \pm 1.0$} & \reb{$3.33 \pm 0.12$} \\
\reb{Foehn}      & \reb{$22.2 \pm 0.4$} & \reb{$1.0 \pm 0.2$}  & \reb{$2.47 \pm 0.05$} \\
\grayrow
\reb{AD-Planner} & \reb{$16.1 \pm 0.2$} & \reb{$1.1 \pm 0.2$}  & \reb{$1.09 \pm 0.06$} \\
\xmidrule{4}
\reb{Racing~\cite{geles2026racing}} & \reb{$47.9 \pm 1.1$} & \reb{$12.9 \pm 0.3$} & \reb{--} \\
\xbottomrule{4}
\end{tabular}
\vspace{-0.4cm}
\end{table}

\reb{\textit{Utilization of the Platform.}
To determine whether the shorter throws result from better utilization of the platform, Table~\ref{tab:agility_metrics} reports the thrust and body rate the quadrotor achieves between throw start and payload release, and the payload speed at release, at the \qty{2.5}{\meter} target.
Our policy reaches over \qty{80}{\percent} of the peak body rate and specific thrust of racing flights with a quadrotor of the same model (Table~\ref{tab:agility_metrics}, last row); both baselines use less than \qty{60}{\percent} of our peak thrust and approximately one tenth of our peak body rate.
This authority goes into a single \qty{0.4}{\second} swing: coordinated translational and rotational motion builds up the payload's momentum and yields a faster payload at release, whereas the baselines ramp up gradually over \qtyrange{1.1}{2.7}{\second}.
The learned throw thus makes fuller use of the platform's authority while maintaining the accuracy reported in Table~\ref{tab:throwing_results}, whereas the baselines leave most of it unused, limited by their simplified models and constraints.}

\subsection{Ablation Studies}
\label{subsec:ablation_studies}
\begin{figure}[!t]
\centering
    \includegraphics[width=0.75\columnwidth]{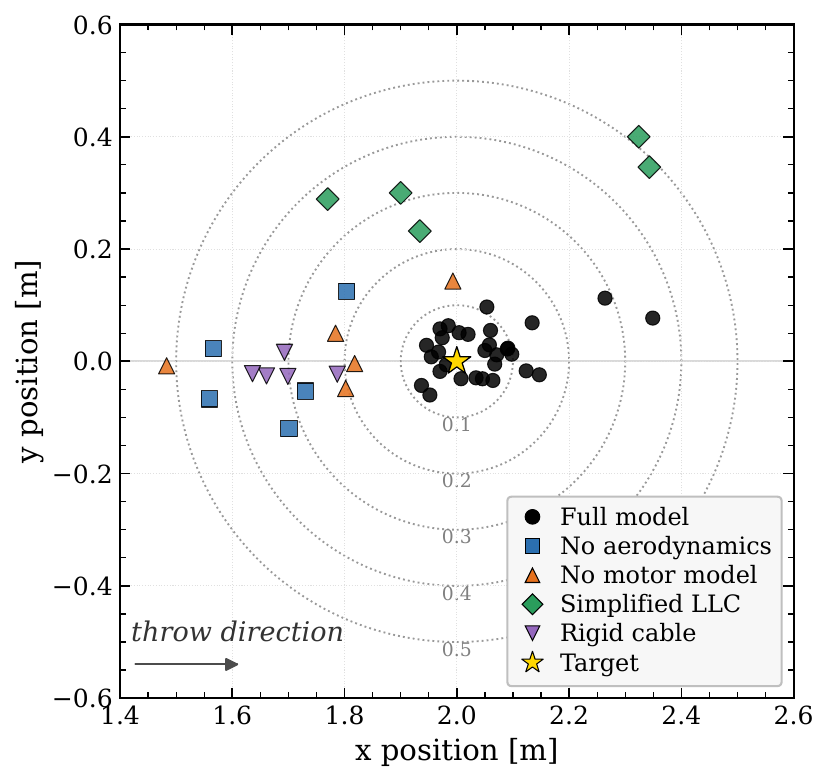}
\caption{Top-down landing scatter at the \qty{2.0}{\meter} target for the full model and the ablations \reb{(thirty throws for the full model, five per ablation)}. \reb{The full model lands closer to the target. The aerodynamics, motor, and rigid-cable ablations undershoot, while the simplified controller overshoots with a lateral bias. The rigid-cable policy retains comparable precision but has a systematic offset.}}
\label{fig:ablation_landing}
\end{figure}

To investigate the effect of each component of the hybrid simulator on sim-to-real transfer~\cite{aljalbout2025reality}, we retrain the policy with a single component of the simulation pipeline removed or simplified at a time, keeping the PPO recipe, reward, and domain randomization identical. 
Each resulting policy is deployed zero-shot on the real platform under the same setup as Sec.~\ref{subsec:throwing_experiments}.
We consider four ablations, each isolating one element of the analytical quadrotor model (Table~\ref{tab:analytical_blocks}) or the articulated payload: (i) \emph{no aerodynamics}, removing the rotor-inflow and drag augmentation block; (ii) \emph{no motor model}, collapsing the first-order motor response to an instantaneous one; (iii) \emph{simplified low-level controller}, replacing the identified Betaflight \ac{CTBR} controller with a simple geometric body-rate controller; and (iv) \emph{rigid cable}, replacing the flexible multi-link tether with a single rigid rod of identical length and mass. As reported in Table~\ref{tab:ablation_results}, removing any single one of these components increases the landing error by $2.9$--$4.4\times$ at the \qty{2.0}{\meter} target while leaving the throw duration essentially unchanged, confirming that each of these components contributes to \rev{accurate} sim-to-real transfer. We ablate the components expected to dominate the transfer gap; the remaining blocks are identified directly from the platform and held fixed, and are not varied here. The corresponding landing scatter (Fig.~\ref{fig:ablation_landing}) makes the failure modes explicit: the aerodynamics, motor, and rigid-cable ablations systematically undershoot the target, whereas the simplified low-level controller overshoots with a large lateral bias.

\begin{table}[!t]
\centering
\caption{Ablation study on hardware at the \qty{2.0}{\meter} target. Each policy is retrained with a single simulation component removed or simplified, then deployed zero-shot under the protocol of Sec.~\ref{subsec:throwing_experiments}.}
\label{tab:ablation_results}
\footnotesize
\setlength{\tabcolsep}{3pt}
\begin{tabular}{@{}lcc@{}}
\xtoprule{3}
\textbf{Configuration} & \textbf{Landing Error [\si{\meter}]} & \textbf{Throw Duration [\si{\second}]} \\
\xmidrule{3}
Full model & \reb{$\mathbf{0.088 \pm 0.070}$} & \reb{$1.142 \pm 0.014$} \\
\xmidrule{3}
\grayrow
No aerodynamics                & $0.342 \pm 0.085$ & $1.069 \pm 0.008$ \\
No motor model                 & $0.254 \pm 0.134$ & $1.109 \pm 0.032$ \\ \grayrow
Simplified low-level controller & $0.386 \pm 0.103$ & $1.115 \pm 0.042$ \\
Rigid cable                    & $0.306 \pm 0.051$ & $1.049 \pm 0.009$ \\
\xbottomrule{3}
\end{tabular}
\end{table}

\subsection[Visual Feature-Based Throwing]{\reb{Visual Feature-Based Throwing}}
\label{subsec:vision_throwing}
\reb{We extend the framework to visual feature-based throwing and evaluate the policy using \ac{HIL}, following~\cite{xing2024bootstrapping, heegIcra2025, pan2025learning}. As in~\cite{heegIcra2025}, virtual visual observations allow us to assess control performance independently of feature detection performance of the perception pipeline. The quadrotor, cable, payload, and release mechanism are physical, while a virtual camera generates the visual features.}
\reb{Motion capture measures the quadrotor pose and payload position. At each policy step, the virtual downward-facing fisheye camera is placed at the measured quadrotor pose and projects a known $\qty{0.25}{\meter}$ target square and four points around the measured payload center into normalized image coordinates. The actor receives the current target keypoints and three frames of payload keypoints, replacing the explicit goal offset and payload-position history. Quadrotor position, attitude, linear and angular velocity estimates, previous action, and attachment flag remain inputs. The critic additionally receives privileged goal and payload states during training only.}

\reb{Deployed zero-shot as in Sec.~\ref{subsec:throwing_experiments}, this policy achieves similar landing accuracy and throw duration to the state-based policy (Table~\ref{tab:vision_results}). Fig.~\ref{fig:vision_sequence} visualizes the projected features. These results demonstrate that the policy can perform accurate throws using visual keypoints for the target and payload, providing a basis for future integration with a real visual perception system.}

\newcommand{\visframe}[1]{%
\begin{tikzpicture}
\node[anchor=south west,inner sep=0] (vimg) {\includegraphics[width=0.40\columnwidth]{#1}};
\begin{scope}[x={(vimg.south east)},y={(vimg.north west)}]
  \draw[->,white,line width=0.6pt] (0.06,0.10) -- (0.27,0.10) node[anchor=west,inner sep=1pt,text=white]{\scriptsize$x$};
  \draw[->,white,line width=0.6pt] (0.06,0.10) -- (0.06,0.34) node[anchor=south,inner sep=1pt,text=white]{\scriptsize$y$};
\end{scope}
\end{tikzpicture}}
\begin{figure}[!t]
\centering
\subfloat[$t=0.00$\,s]{\visframe{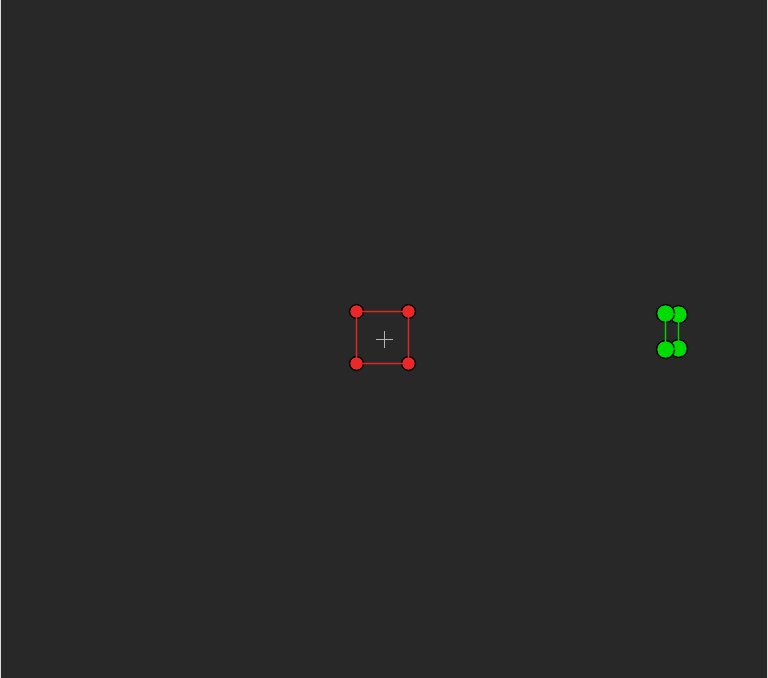}}\hfil
\subfloat[$t=0.14$\,s]{\visframe{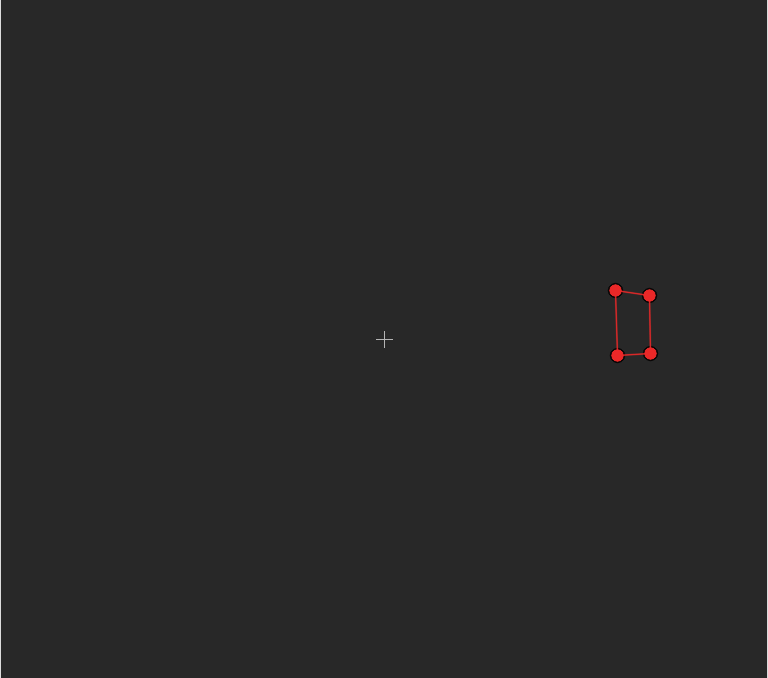}}\\
\subfloat[$t=0.40$\,s]{\visframe{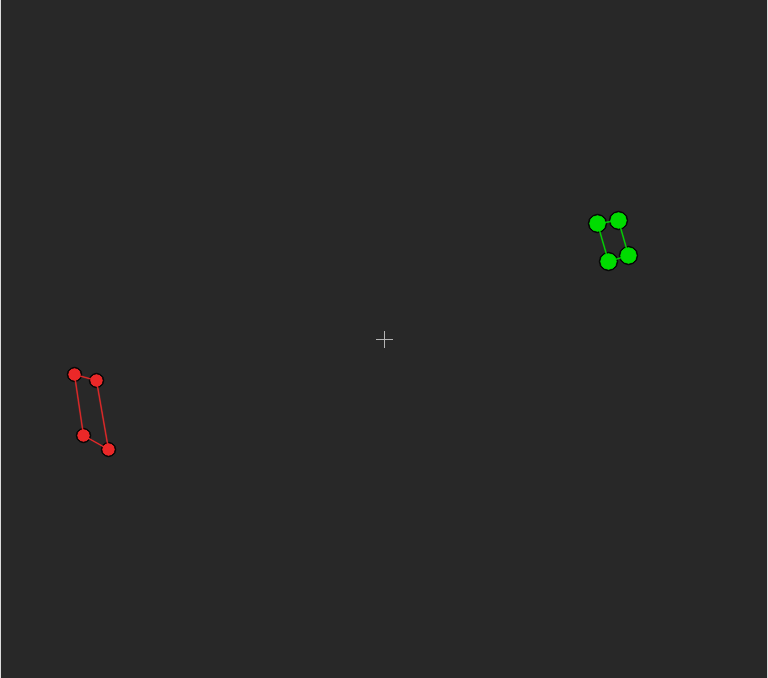}}\hfil
\subfloat[$t=1.10$\,s]{\visframe{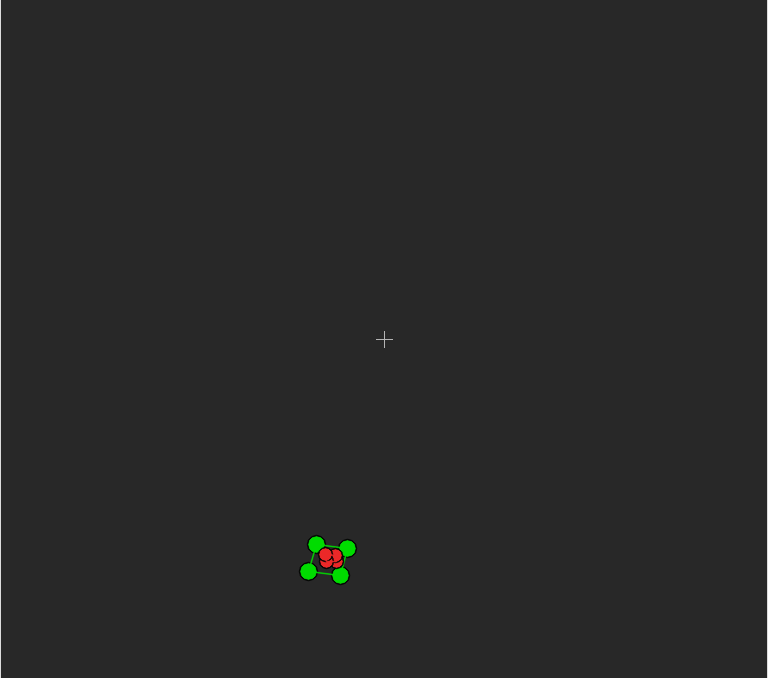}}
\caption{\reb{Visual feature-based throwing experiment (HIL): visualization of virtual visual observations generated from motion capture during a physical throw to the \qty{2.0}{\meter} target.} \textcolor{green}{Green}: goal keypoints; \textcolor{red}{red}: payload keypoints. The white arrows indicate the body-frame axes. Times are relative to policy start: (a) policy start; (b) quadrotor pitches down to accelerate forward, driving the payload forward in the image; (c) release; (d) landing.}
\label{fig:vision_sequence}
\vspace{-0.3cm}
\end{figure}


\begin{table}[!t]
\centering
\caption{\reb{Visual feature-based policy versus the state-based policy on hardware. Ten throws per target for the visual feature-based policy, thirty for the state-based policy (Table~\ref{tab:throwing_results}).}}
\label{tab:vision_results}
\footnotesize
\setlength{\tabcolsep}{5pt}
\begin{tabular}{@{}llcc@{}}
\xtoprule{4}
\textbf{Target} & \textbf{Observation} & \textbf{Landing Error [\si{\meter}]} & \textbf{Throw Duration [\si{\second}]} \\
\xmidrule{4}
\multirow{2}{*}{\qty{1.5}{\meter}} & \cellcolor{gray!10}Explicit state & \cellcolor{gray!10}\reb{$0.079 \pm 0.030$} & \cellcolor{gray!10}\reb{$1.084 \pm 0.013$} \\
                                   & \reb{Visual features} & $0.101 \pm 0.058$ & $1.051 \pm 0.011$ \\
\xmidrule{4}
\multirow{2}{*}{\qty{2.0}{\meter}} & \cellcolor{gray!10}Explicit state & \cellcolor{gray!10}\reb{$0.088 \pm 0.070$} & \cellcolor{gray!10}\reb{$1.142 \pm 0.014$} \\
                                   & \reb{Visual features} & $0.101 \pm 0.106$ & $1.125 \pm 0.029$ \\
\xbottomrule{4}
\end{tabular}
\end{table}

%% file: sections/conclusion.tex
\section{Conclusion}
We presented an RL-based approach for agile and accurate throwing of a cable-suspended payload from a quadrotor. \rev{It relies on} a hybrid simulator that couples an identified analytical quadrotor model with an external physics solver for the rope and payload through a single wrench channel, allowing a policy trained entirely in simulation to throw accurately in the real world, zero-shot. \reb{Real-world experiments against two state-of-the-art baselines show that optimizing the throw end-to-end yields substantially more agile throws \rev{with mean landing errors below \qty{12}{\centi\meter}}, avoiding the plan-then-track bottleneck and the conservative constraints of the model-based pipelines}, and \rev{our ablation studies show that each tested simulator component is necessary for accurate transfer}. These results are obtained under three simplifying assumptions that also outline future work: targets are static on the ground plane, the payload is fixed in mass, and the post-release trajectory is predicted using a drag-free ballistic model. \rev{Within this scope, the policy transfers zero-shot up to the out-of-distribution \qty{2.5}{\meter} target and, in simulation, retains its nominal accuracy for payload-mass errors within $\pm10\%$ (Fig.~\ref{fig:sim_robustness}). We expect real-world data, e.g., for fine-tuning~\cite{pan2025learning, xing2026continual}, to become necessary beyond these conditions: for targets farther than \qty{2.5}{\meter}, where the simulated error grows rapidly, for larger payload-mass errors, and for payloads subject to significant drag.}

%% file: main.bbl
\begin{thebibliography}{10}
\providecommand{\url}[1]{#1}
\csname url@samestyle\endcsname
\providecommand{\newblock}{\relax}
\providecommand{\bibinfo}[2]{#2}
\providecommand{\BIBentrySTDinterwordspacing}{\spaceskip=0pt\relax}
\providecommand{\BIBentryALTinterwordstretchfactor}{4}
\providecommand{\BIBentryALTinterwordspacing}{\spaceskip=\fontdimen2\font plus
\BIBentryALTinterwordstretchfactor\fontdimen3\font minus
  \fontdimen4\font\relax}
\providecommand{\BIBforeignlanguage}[2]{{%
\expandafter\ifx\csname l@#1\endcsname\relax
\typeout{** WARNING: IEEEtran.bst: No hyphenation pattern has been}%
\typeout{** loaded for the language `#1'. Using the pattern for}%
\typeout{** the default language instead.}%
\else
\language=\csname l@#1\endcsname
\fi
#2}}
\providecommand{\BIBdecl}{\relax}
\BIBdecl

\bibitem{Foehn_Falanga_Kuppuswamy_Tedrake_Scaramuzza_2017}
P.~Foehn, D.~Falanga, N.~Kuppuswamy, R.~Tedrake, and D.~Scaramuzza, ``Fast
  trajectory optimization for agile quadrotor maneuvers with a cable-suspended
  payload,'' in \emph{Proc. Robot.: Sci. Syst. (RSS)}, Jul. 2017.

\bibitem{Cao_Fang_Liang_2025}
R.~Cao, Y.~Fang, and X.~Liang, ``Time-optimal trajectory planning with clearly
  defined initial guess for aerial suspended payload throwing,'' \emph{IEEE
  Trans. Autom. Sci. Eng.}, 2025.

\bibitem{AD_planner}
H.~Li, Y.~Xu, C.~Feng, B.~Zhou, and S.~Shen, ``{AD-Planner}: Adaptive in-flight
  delivery using a quadrotor with a suspended payload,'' \emph{IEEE Trans.
  Robot.}, vol.~42, pp. 3304--3323, 2026.

\bibitem{pcmpc2021}
G.~Li, A.~Tunchez, and G.~Loianno, ``Pcmpc: Perception-constrained model
  predictive control for quadrotors with suspended loads using a single camera
  and imu,'' in \emph{Proc. IEEE Int. Conf. Robot. Autom. (ICRA)}, 2021, pp.
  2012--2018.

\bibitem{zeng2025decentralizedaerialmanipulationcablesuspended}
J.~Zeng, A.~M. Gimenez, E.~Vinitsky, J.~Alonso-Mora, and S.~Sun,
  ``Decentralized aerial manipulation of a cable-suspended load using
  multi-agent reinforcement learning,'' in \emph{Proc. Conf. Robot Learn.
  (CoRL)}, 2025.

\bibitem{flare2026}
D.~Cao, J.~Zhou, X.~Wang, and S.~Li, ``{FLARE}: Agile flights for quadrotor
  cable-suspended payload system via reinforcement learning,'' \emph{IEEE
  Robot. Autom. Lett.}, vol.~PP, pp. 1--8, 2026.

\bibitem{xing2024bootstrapping}
J.~Xing, A.~Romero, L.~Bauersfeld, and D.~Scaramuzza, ``Bootstrapping
  reinforcement learning with imitation for vision-based agile flight,'' in
  \emph{Proc. Conf. Robot Learn. (CoRL)}, 2024.

\bibitem{heegIcra2025}
J.~Heeg, Y.~Song, and D.~Scaramuzza, ``Learning quadrotor control from visual
  features using differentiable simulation,'' in \emph{Proc. IEEE Int. Conf.
  Robot. Autom. (ICRA)}, 2025.

\bibitem{pan2025learning}
J.~Pan, J.~Xing, R.~Reiter, Y.~Zhai, E.~Aljalbout, and D.~Scaramuzza,
  ``Learning on the fly: Rapid policy adaptation via differentiable
  simulation,'' \emph{IEEE Robot. Autom. Lett.}, 2026.

\bibitem{6760219}
K.~Sreenath, T.~Lee, and V.~Kumar, ``Geometric control and differential
  flatness of a quadrotor uav with a cable-suspended load,'' in \emph{Proc.
  IEEE Conf. Decis. Control (CDC)}, 2013, pp. 2269--2274.

\bibitem{panetsos2023event}
F.~Panetsos, G.~Karras, and K.~Kyriakopoulos, ``Aerial transportation of
  cable-suspended loads with an event camera,'' \emph{IEEE Robot. Autom.
  Lett.}, vol.~PP, pp. 1--8, 2023.

\bibitem{li2023autotrans}
H.~Li, H.~Wang, C.~Feng, F.~Gao, B.~Zhou, and S.~Shen, ``Autotrans: A complete
  planning and control framework for autonomous uav payload transportation,''
  \emph{IEEE Robot. Autom. Lett.}, 2023.

\bibitem{10478625}
H.~Wang, H.~Li, B.~Zhou, F.~Gao, and S.~Shen, ``Impact-aware planning and
  control for aerial robots with suspended payloads,'' \emph{IEEE Trans.
  Robot.}, vol.~40, pp. 2478--2497, 2024.

\bibitem{sun2025agilecooperative}
S.~Sun, X.~Wang, D.~Sanalitro, A.~Franchi, M.~Tognon, and J.~Alonso-Mora,
  ``Agile and cooperative aerial manipulation of a cable-suspended load,''
  \emph{Sci. Robot.}, vol.~10, no. 107, p. eadu8015, 2025.

\bibitem{recalde2025eshpc}
L.~Recalde, M.~Sarvaiya, G.~Loianno, and G.~Li, ``Es-hpc-mpc: Exponentially
  stable hybrid perception constrained mpc for quadrotor with suspended
  payloads,'' \emph{IEEE Robot. Autom. Lett.}, 2025.

\bibitem{10328685}
G.~Li, X.~Liu, and G.~Loianno, ``Rotortm: A flexible simulator for aerial
  transportation and manipulation,'' \emph{IEEE Trans. Robot.}, vol.~40, pp.
  831--850, 2024.

\bibitem{rapuano2026nonlinear}
A.~Rapuano, Y.~Shen, F.~Califano, C.~Gabellieri, and A.~Franchi, ``Nonlinear
  predictive control of the continuum and hybrid dynamics of a suspended
  deformable cable for aerial pick and place,'' in \emph{Proc. IEEE Int. Conf.
  Robot. Autom. (ICRA)}, 2026.

\bibitem{Belkhale_2021}
S.~Belkhale, R.~Li, G.~Kahn, R.~McAllister, R.~Calandra, and S.~Levine,
  ``Model-based meta-reinforcement learning for flight with suspended
  payloads,'' \emph{IEEE Robot. Autom. Lett.}, vol.~6, no.~2, 2021.

\bibitem{xu2024omnidrones}
B.~Xu, F.~Gao, C.~Yu, R.~Zhang, Y.~Wu, and Y.~Wang, ``{OmniDrones}: An
  efficient and flexible platform for reinforcement learning in drone
  control,'' \emph{IEEE Robot. Autom. Lett.}, vol.~9, no.~3, pp. 2838--2845,
  2024.

\bibitem{mittal2025isaaclab}
M.~Mittal \emph{et~al.}, ``Isaac lab: A gpu-accelerated simulation framework
  for multi-modal robot learning,'' \emph{arXiv preprint arXiv:2511.04831},
  2025.

\bibitem{Macklin_Warp_A_High-performance_2022}
\BIBentryALTinterwordspacing
M.~Macklin, ``{Warp: A High-performance Python Framework for GPU Simulation and
  Graphics},'' Mar. 2022. [Online]. Available:
  \url{https://github.com/NVIDIA/warp}
\BIBentrySTDinterwordspacing

\bibitem{nvidia_physx}
\BIBentryALTinterwordspacing
{NVIDIA}, ``{NVIDIA PhysX SDK},'' 2026. [Online]. Available:
  \url{https://github.com/NVIDIA-Omniverse/PhysX}
\BIBentrySTDinterwordspacing

\bibitem{foehn2022agilicious}
P.~Foehn \emph{et~al.}, ``Agilicious: Open-source and open-hardware agile
  quadrotor for vision-based flight,'' \emph{Sci. Robot.}, vol.~7, 06 2022.

\bibitem{song2023rl_vs_oc}
Y.~Song, A.~Romero, M.~Müller, V.~Koltun, and D.~Scaramuzza, ``Reaching the
  limit in autonomous racing: Optimal control versus reinforcement learning,''
  \emph{Sci. Robot.}, vol.~8, no.~82, p. eadg1462, 2023.

\bibitem{xing2024multi}
J.~Xing, I.~Geles, Y.~Song, E.~Aljalbout, and D.~Scaramuzza, ``Multi-task
  reinforcement learning for quadrotors,'' \emph{IEEE Robot. Autom. Lett.},
  vol.~10, no.~3, pp. 2112--2119, 2024.

\bibitem{Zhai2025PAMPPIPM}
Y.~Zhai, R.~Reiter, and D.~Scaramuzza, ``Pa-mppi: Perception-aware model
  predictive path integral control for quadrotor navigation in unknown
  environments,'' \emph{IEEE Robot. Autom. Lett.}, 2025.

\bibitem{falanga2020obstacle}
D.~Falanga, K.~Kleber, and D.~Scaramuzza, ``Dynamic obstacle avoidance for
  quadrotors with event cameras,'' \emph{Sci. Robot.}, vol.~5, no.~40, p.
  eaaz9712, 2020.

\bibitem{merk2026learningacrobaticflightpreferences}
C.~Merk, I.~Geles, J.~Xing, A.~Romero, G.~Ramponi, and D.~Scaramuzza,
  ``Learning acrobatic flight from preferences,'' \emph{arXiv preprint}, 2026.

\bibitem{gao2025aerobatics}
M.~Wang, Q.~Wang, Z.~Wang, Y.~Gao, J.~Wang, C.~Cui, Y.~Li, Z.~Ding, K.~Wang,
  C.~Xu, and F.~Gao, ``Unlocking aerobatic potential of quadcopters: Autonomous
  freestyle flight generation and execution,'' \emph{Sci. Robot.}, vol.~10, no.
  101, p. eadp9905, 2025.

\bibitem{schulman2017proximal}
J.~Schulman, F.~Wolski, P.~Dhariwal, A.~Radford, and O.~Klimov, ``Proximal
  policy optimization algorithms,'' \emph{arXiv preprint}, 2017.

\bibitem{zhou2019continuity}
Y.~Zhou, C.~Barnes, J.~Lu, J.~Yang, and H.~Li, ``On the continuity of rotation
  representations in neural networks,'' in \emph{Proc. IEEE/CVF Conf. Comput.
  Vis. Pattern Recognit. (CVPR)}, 2019, pp. 5745--5753.

\bibitem{newton2025engine}
\BIBentryALTinterwordspacing
{Newton Physics Contributors}, ``Newton: An open-source, {GPU}-accelerated
  physics simulation engine,'' 2025. [Online]. Available:
  \url{https://github.com/newton-physics/newton}
\BIBentrySTDinterwordspacing

\bibitem{geles2026racing}
I.~Geles, L.~Bauersfeld, M.~Wulfmeier, and D.~Scaramuzza, ``Superhuman safe and
  agile racing through multi-agent reinforcement learning,'' \emph{arXiv
  preprint arXiv:2605.22748}, 2026.

\bibitem{aljalbout2025reality}
E.~Aljalbout, J.~Xing, A.~Romero, I.~Akinola, C.~R. Garrett, E.~Heiden,
  A.~Gupta, T.~Hermans, Y.~Narang, D.~Fox \emph{et~al.}, ``The reality gap in
  robotics: Challenges, solutions, and best practices,'' \emph{Annu. Rev.
  Control Robot. Auton. Syst.}, vol.~9, 2025.

\bibitem{xing2026continual}
J.~Xing, Z.~Zhu, Y.~Ren, I.~Geles, Y.~Zhai, R.~Reiter, and D.~Scaramuzza,
  ``Continual robot policy learning via variational neural dynamics,''
  \emph{arXiv preprint arXiv:2606.27353}, 2026.

\end{thebibliography}
